%% file: main.tex
\pdfoutput=1
\documentclass{article}

\usepackage[siunitx]{circuitikz}
\usetikzlibrary{positioning}
\usepackage{tikz}
\usepackage{amsmath}
\usepackage{graphicx}
\usepackage{etoolbox}
\usepackage{changepage}

\usepackage{xurl} %

\usepackage{appendix}

\makeatletter
\newcommand{\appsection}[1]{%
  \refstepcounter{section}%
  \section*{\normalsize Appendix~\thesection.\ #1}%
  \addcontentsline{toc}{section}{Appendix~\thesection.\ #1}%
}
\newcommand{\appsubsection}[1]{%
  \refstepcounter{subsection}%
  \subsection*{\normalsize Appendix~\thesection.\arabic{subsection}\ #1}%
  \addcontentsline{toc}{subsection}{Appendix~\thesection.\arabic{subsection}\ #1}%
}
\makeatother

\usepackage{amssymb}
\usepackage{amsmath}
\usepackage{graphicx}
\usepackage{booktabs}
\usepackage{makecell}
\usepackage{tabularx}
\usepackage{array}
\usepackage{float}
\usepackage{pifont}
\usepackage{subcaption}   %
\usepackage{microtype}
\usepackage[nolist]{acronym}
\newcolumntype{C}{>{\centering\arraybackslash}X}

\usepackage[colorlinks=true,linkcolor=cyan,citecolor=cyan,urlcolor=cyan]{hyperref}
\usepackage[capitalise]{cleveref}

\usepackage{spconf}
\makeatletter
\renewcommand{\title}[1]{\gdef\@title{#1}}
\makeatother
\usepackage{graphicx}
\usepackage{mathtools} %
\usepackage{amssymb,amsfonts}

\usepackage[ruled,vlined]{algorithm2e}

\usepackage{booktabs}
\usepackage{multirow}
\usepackage{tabularx,array}

\usepackage{siunitx}
\usepackage{bm}
\usepackage{enumitem}
\usepackage{xcolor}
\usepackage{cite}

\usepackage{hyperref}

\DeclareSIUnit{\var}{VAr}
\DeclareSIUnit{\voltampereReactive}{VAr}
\DeclareSIUnit{\ohm}{\text{\ensuremath{\Omega}}}
\newcolumntype{Y}{>{\raggedleft\arraybackslash}X}

\def\BibTeX{{\rm B\kern-.05em{\sc i\kern-.025em b}\kern-.08em
    T\kern-.1667em\lower.7ex\hbox{E}\kern-.125emX}}

\usepackage{placeins}
\graphicspath{{img/}}

\begin{document}
\input{acronyms}
\title{DRetHTR: Linear-Time Decoder-Only Retentive Network for Efficient Handwritten Text Recognition}

\name{%
\begin{tabular}{@{}c@{}}
Changhun Kim$^{1}$%
\thanks{Corresponding author: \texttt{changhun.kim@fau.de}}%
\thanks{Code: \url{https://github.com/Kimchangheon/DRetHTR}}%
, Martin Mayr$^{1}$, Thomas Gorges$^{1}$, Fei Wu$^{1}$,
Mathias Seuret$^{1}$, \\Andreas Maier$^{1}$, Vincent Christlein$^{1}$
\end{tabular}
}

\address{
$^{1}$ Pattern Recognition Lab, Friedrich-Alexander-Universität Erlangen-Nürnberg, Germany
}

\ninept
\maketitle

\begin{abstract}
State-of-the-art handwritten text recognition (HTR) systems commonly use Transformers, whose growing key-value (KV) cache makes decoding slow and memory-intensive. We introduce DRetHTR, a decoder-only model built on Retentive Networks (RetNet). Compared to an equally sized decoder-only Transformer baseline, DRetHTR delivers 1.6–1.9$\times$ faster inference with 38–42\% less memory usage, without loss of accuracy. By replacing softmax attention with softmax-free retention and injecting multi-scale sequential priors, DRetHTR avoids a growing KV cache: decoding is linear in output length in both time and memory. To recover the local-to-global inductive bias of attention, we propose layer-wise gamma scaling, which progressively enlarges the effective retention horizon in deeper layers. This encourages early layers to model short-range dependencies and later layers to capture broader context, mitigating the flexibility gap introduced by removing softmax. Consequently, DRetHTR achieves best reported test character error rates of 2.26\% on IAM-A and 1.81\% on RIMES, while remaining competitive on READ-2016 with 4.21\%. Evaluation across ten handwriting datasets and nine scene-text benchmarks further confirms the robustness of the model across diverse languages, scripts, writing styles, and imaging conditions. This demonstrates that decoder-only RetNet enables Transformer-level HTR accuracy with substantially improved decoding speed and memory efficiency.

\end{abstract}


\section{Introduction}
\label{sec:intro}

Digitized documents permeate healthcare, insurance, government, and finance, yet a vast body of handwritten content remains locked in archives and libraries worldwide. \ac{htr} addresses these challenges by teaching machines to decode handwriting: platforms such as Transkribus~\cite{Nockels2022} have demonstrated the practical impact of large-scale transcription for historians, genealogists, and the digital humanities, accelerating scholarship and public access to archival records. Beyond cultural heritage, modern administrative workflows across sectors have also been automated by \ac{htr}, saving enormous manual labor and improving productivity~\cite{blueprism_handwriting_ocr}.

HTR is a pattern recognition problem over visual sequences: the system must map an input image to a character sequence while leveraging linguistic regularities in context.
Initial approaches such as \acp{hmm} ~\cite{HMM_HTR} transitioned to more complex architectures like \ac{bilstm}. However, distinguishing between similar characters is still challenging without context knowledge, which is usually handled by external language models~\cite{statisticallanguage}.

Line-based HTR commonly combines \acp{cnn} and \acp{rnn}~\cite{CNNBiLSTMCTC}, trained with the \ac{ctc} loss~\cite{CTC}.
This enables alignment between input image and output character sequences without predefined character segmentation.
However, limitations such as enforcing strict input-output sequence alignment and limiting the output lengths remain.
This motivates the usage of sequence-to-sequence architectures, especially Transformers~\cite{Transformer}, allowing for unrestricted sequence length and the use of contextual cues.

\begin{figure}[t]
  \centering
  \includegraphics[width=0.9\linewidth]{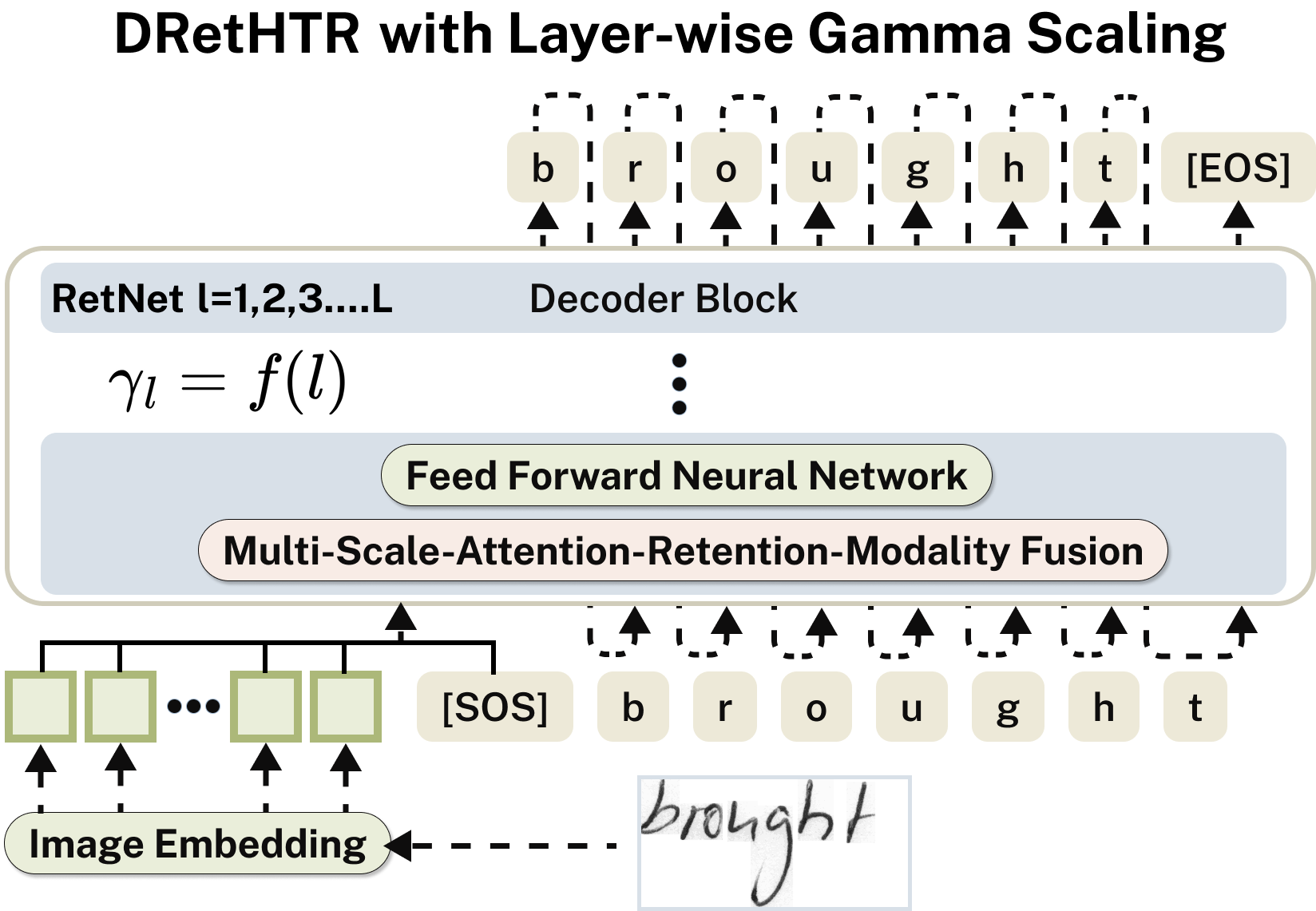}
  \caption{Decoder-only RetNet architecture that fuses the image and text in the Decoder}
  \label{fig:intro_image}
\end{figure}

DTrOCR~\cite{DTrOCR} was introduced as an effective decoder-only architecture for \ac{htr}, surpassing previous models by leveraging the pre-trained GPT-2~\cite{gpt2}, which provides robust natural language understanding.
However, Transformer self-attention must form (and typically store) an $N\times N$ attention map for a sequence of length $N$, making long sequences costly.
In contrast, \acp{rnn} decode with constant per-step cost and long-sequence memory that grows only linearly with $N$, but they lack Transformer-style parallel training.

\acp{retnet} address this trade-off by combining parallel training with a softmax-free \emph{retention} mechanism that admits a recurrent form at inference~\cite{retnet}.
This enables constant-cost decoding per step with linear long-sequence memory, while maintaining competitive performance, making RetNet an attractive foundation for efficient HTR.

To harness these advantages for \ac{htr}, we introduce DRetHTR, a decoder-only \ac{retnet} tailored to line-level handwriting that achieves linear-time, linear-memory decoding. At the decoder input, DRetHTR employs an \emph{\ac{armf}} layer that aligns image and text tokens and provides a \emph{modality-partitioned interface}: image-key interactions use standard attention for alignment, while text-key interactions use \emph{causal retention} with lower-triangular decay matrix D. To recover the local-to-global inductive bias typically provided by softmax attention, we propose \emph{layer-wise} $\gamma$ \emph{scaling} for retention: shallow layers use smaller decay factors to emphasize local dependencies, while deeper layers increase the decay to aggregate broader context. Our ablation study confirms that this schedule matches a Transformer’s CER while preserving linear-time, linear-memory decoding. Under a unified implementation and evaluation framework, we assess DRetHTR on ten line-level handwriting datasets spanning modern and historical documents, Latin and Chinese scripts, multiple languages, and diverse writing styles, including IAM~\cite{Marti2002IAM}, RIMES~\cite{RIMES}, READ-2016~\cite{READ2016}, and SCUT-HCCDoc~\cite{Zhang2020SCUTHCCDoc}. DRetHTR delivers competitive accuracy while achieving \mbox{1.6–1.9$\times$} faster inference and \mbox{38–42\,\%} lower memory usage. We additionally evaluate its transferability on nine scene-text benchmarks.

The paper is structured as follows. \Cref{sec:formatting} reviews related work on HTR architectures. \Cref{sec:RetNet} provides background on RetNet and its variants. \Cref{sec:method} presents DRetHTR, including \ac{armf} and layer-wise $\gamma$ scaling. \Cref{sec:experiments} reports quantitative and qualitative results, including ablations. \Cref{sec:limit} and \Cref{sec:future} discuss limitations and future work, and \Cref{sec:conclude} concludes.

\section{Related Work}
\label{sec:formatting}
In the early days of \ac{htr}, classical machine learning models such as \ac{svm} and \ac{knn} were widely used.
These methods relied on hand-crafted features extracted from geometric, statistical, or polar-transformed representations of text images.
Similarly, Farah~\textit{et al.}~\cite{KNN_HTR} use a parallel combination of \ac{knn}, fuzzy \ac{knn}, and neural networks to recognize Arabic words with a 96\,\% accuracy rate. Despite good single-character performance, these methods require tedious segmentation of line images into individual characters or strokes.

\acp{hmm} use the sliding window technique to extract a sequence of feature vectors from an entire line of text without needing pre-segmented data. Plotz~\textit{et al.}~\cite{HMM_HTR} emphasized that \acp{hmm} perform implicit alignment between input features and target sequences using the Viterbi decoding. The dominance of \acp{rnn} in handwriting recognition was enabled by the introduction of the \ac{ctc} loss function~\cite{CTC}. Moreover, additional mechanisms such as MDLSTMs~\cite{MDLSTM} and gated convolutional recurrent neural networks~\cite{Bluche2017} further improved \ac{htr} performance.

Recent developments introduced Transformer-based approaches, significantly enhancing parallel processing, scalability, and global dependency modeling. Notable methods include NRTR~\cite{NRTR}, fully Transformer-based architectures by Kang~\textit{et al.}~\cite{Kang}, and hybrid Transformer-\ac{ctc} models~\cite{Diaz2021RethinkingTL,htrvt}. For diplomatic transcription in low-data environments, Mayr~\textit{et al.}~\cite{Mayr2025DiplomaticHTR} made practical modifications to a Transformer encoder-decoder to bypass the length constraints of CTC. In parallel, several works report that pairing a Transformer decoder with a CTC auxiliary branch on the encoder—or leveraging CTC during decoding—improves data efficiency and recognition accuracy~\cite{barrere2022light,Barrere2024FewDataHTR,Wick2022CTCPrefixRescoring}.

Large-scale pre-trained families, notably TrOCR and DTrOCR, leverage foundational Transformer language models and achieve state-of-the-art accuracy~\cite{TrOCR,DTrOCR}. TrOCR is an end-to-end encoder–decoder that couples a ViT/BEiT-style image encoder with a RoBERTa-initialized text Transformer decoder~\cite{TrOCR}. By contrast, DTrOCR pushes toward a decoder-only design: images are patch-tokenized and fed to a Transformer decoder initialized from GPT-2, which then autoregressively generates the transcription~\cite{DTrOCR,gpt2}. This simplification exploits strong language priors and removes encoder–decoder cross-attention. Nevertheless, it retains the sequence-length-dependent computational and memory costs associated with softmax self-attention and the growing \ac{kv} cache during autoregressive decoding.

Beyond improving recognition accuracy through larger architectures and large-scale pre-training, recent work has increasingly focused on reducing decoding cost by redesigning the generation process itself. Meta-DAN uses windowed queries and confidence-based multi-token prediction for page-level \ac{htr}, while DiffusionSTR iteratively refines a fixed-length noisy token sequence for scene-text recognition~\cite{MetaDAN2026,DiffusionSTR2023}. Both approaches reduce the number of strictly sequential decoding steps.

DRetHTR addresses a complementary efficiency bottleneck by preserving autoregressive beam search and explicit causal language factorization while replacing the sequence-length-dependent text \ac{kv} cache with fixed-size recurrent retention states. Collectively, these studies demonstrate that decoding strategy has become a central component of efficient text recognition.

\section{Retentive Networks -- Background}
\label{sec:RetNet}
Transformer's self-attention enables fully parallel training; during autoregressive decoding, however, standard implementations maintain a \ac{kv} cache that grows linearly with generated length, motivating research on effective KV cache management architectures~\cite{survey_LLM_KV}. \ac{retnet} replaces the growing \ac{kv} cache with a compact recurrent state. Its retention mechanism bridges attention and recurrence: training stays fully parallel, while inference runs as a constant-time recurrence. Retention pursues the same effect as self-attention by circumventing the softmax operation, which prohibits the compatible recurrent representation of self-attention, through the use of a temporal prior decay matrix.

\paragraph*{From self-attention to retention}
Let $X\in\mathbb{R}^{N\times d_{\text{model}}}$ be a length-$N$ sequence. A Transformer maps $X$ to
$Q=XW_Q,\ K=XW_K,\ V=XW_V$ with $W_Q,W_K,W_V\in\mathbb{R}^{d_{\text{model}}\times d_{\text{model}}}$ and computes
$\mathrm{Attn}(X)=\operatorname{softmax}\Bigl(\frac{QK^\top}{\sqrt{d_k}}\Bigr)\,V$.

\paragraph*{Retention: parallel training, recurrent inference}
In the \emph{parallel (training) form}, retention augments $Q$ and $K$ with per-position complex phases
\begin{equation}
Q=(XW_Q)\odot\Theta,\quad
K=(XW_K)\odot\bar{\Theta},\quad
V=XW_V.
\label{eq:qkv-theta}
\end{equation}
where $\odot$ is element-wise multiplication, $\bar{\Theta}$ is the complex conjugate of $\Theta$, and the $n$-th row of $\Theta$ is $\Theta_n=e^{\,\mathrm{i}\,n\theta}$ for position index $n\in\{1,\dots,N\}$.
A lower-triangular decay matrix $D\in\mathbb{R}^{N\times N}$ applies a causal, exponential prior
\begin{equation}
D_{nm}=\begin{cases}
\gamma^{\,n-m}, & n \ge m,\\[2pt]
0, & n < m,
\end{cases}
\label{eq:retention-D}
\end{equation}
with $n$ the query index, $m$ the key index, and $\gamma\in(0,1)$ the decay factor. The parallel retention is then
\begin{equation}
\operatorname{Retention}(X) = (QK^\top \odot D)\,V.
\label{eq:retention-matrix}
\end{equation}
Intuitively, $D$ plays the role of a softmax-free, hand-crafted inductive bias that emphasizes recent context while remaining fully parallelizable for training over $n=1,\ldots, N$.

In the \emph{recurrent (inference) form}, the same operator collapses to a constant-time update that no longer needs a growing KV cache. Defining a state $S_n\in\mathbb{C}^{d_{model}\times d_{model}}$ at step $n$,
\begin{equation}
S_n = \gamma S_{n-1} + k_n^\top v_n, \qquad
\operatorname{Retention}(x_n) = q_n S_n.
\label{eq:retention-recursion}
\end{equation}
where $q_n,k_n,v_n$ are the $n$-th rows of $Q,K,V$.   This yields $O(1)$ per-token computation and $O(N)$ total memory for a length-$N$ sequence, while Transformers require $O(N)$ per-token time from the growing KV cache and $O(N^2)$ memory overall. A summary of parallelism, inference cost, and memory is shown in Tables~\ref{tab:model_comparison} and \ref{tab:computations}. Compared with a KV-cached Transformer, RetNet’s recurrent form is more effective when the model width $d$ is smaller than the sequence length $N$.

\begin{table}[ht]
\centering
\footnotesize %
\setlength{\tabcolsep}{4.3pt} %
\caption{Comparison of model architectures by parallelization, inference cost, memory complexity, and performance.}
\vspace{3pt}
\label{tab:model_comparison}
\begin{tabular}{lcccc}
\toprule
\textbf{Architecture} & \textbf{\makecell{Training \\ parallelization}} & \textbf{\makecell{Inference \\ cost}} & \textbf{\makecell{Long-seq \\ memory}} & \textbf{Performance} \\
\midrule
Transformer & \checkmark & $\mathcal{O}(N)$ & $\mathcal{O}(N^{2})$ & \checkmark\checkmark \\
RetNet      & \checkmark & $\mathcal{O}(1)$ & $\mathcal{O}(N)$     & \checkmark\checkmark \\
RNN         & \ding{55}  & $\mathcal{O}(1)$ & $\mathcal{O}(N)$     & \checkmark \\
\bottomrule
\end{tabular}
\end{table}

\begin{table}[t]
\centering
\footnotesize
\caption{Computational costs of different inference forms. RetNet's recurrent form is more effective than the KV-cached Transformer when model dimension $d$ is less than sequence length $n$; the derivation is given in Appendix~\ref{supp:Asymptotic}.}
\vspace{3pt}
\setlength{\tabcolsep}{3pt}
\begin{tabularx}{\linewidth}{@{}lCC@{}}
\toprule
\textbf{Form} & \textbf{Total Computations} & \makecell{\textbf{Asymptotic Complexity}} \\
\midrule
Vanilla Form         & $2n^2 d + n^2 - 1 + n(d - 1)$ & $\mathcal{O}(n^2 d)$ \\
Key-Value Cached     & $2dn + 2(n - 1)$              & $\mathcal{O}(nd)$ \\
Recurrent Form       & $2d^2 + d - 1$                & $\mathcal{O}(d^2)$ \\
\bottomrule
\end{tabularx}
\label{tab:computations}
\end{table}

\paragraph*{Data-dependent decay (Gated Retention)}
Gated Retention (gRet)~\cite{yoco} improves upon  RetNet~\cite{retnet} by introducing a data-dependent gating mechanism instead of fixed gamma values over the different data samples.
Based on the input data, the model can automatically learn how the influence of preceding tokens decays as their distance increases, while preserving a recurrent representation.
In gRet, \(\gamma\) and \(D\) are defined as follows:
\begin{equation}
\gamma=\operatorname{sigmoid}\!\big(X W_{\gamma}\big)^{1/\tau},\qquad
D_{nm}=
\begin{cases}
\displaystyle \prod_{i=m+1}^{n}\gamma_i, & n \ge m,\\[4pt]
0, & n < m,
\end{cases}
\label{eq:gamma-and-D}
\end{equation}

Here, the temperature \(\tau\) controls decay sharpness by pushing \(\gamma_i\) toward 1: larger \(\tau\) slows decay and lengthens memory, whereas smaller \(\tau\) steepens decay and emphasizes local dependencies; following YOCO~\cite{yoco} we use \(\tau=16\). As in RetNet, a sigmoid value \(\gamma\in(0,1)\) is applied iteratively as a product over the intermediate positions between two tokens, determined by the distance \(|n-m|\).
This means that the closer the tokens are to each other, the stronger their dependency; conversely, the farther apart they are, the weaker the dependency becomes.

\paragraph*{Bi-directional retention for vision}
RMT~\cite{rmt} incorporates spatial priors into Vision Transformers. RMT builds upon the original retention mechanism by first adapting it to a bi-directional form and subsequently enhancing it with a Manhattan distance-based spatial decay. BiRetention becomes suitable for images as it moves from unidirectional to bidirectional decay. BiRetention is expressed as:

\begin{equation}
\operatorname{BiRetention}(X)=\big(QK^{\top}\odot D^{\mathrm{Bi}}\big)\,V,\qquad
D^{\mathrm{Bi}}_{nm}=\gamma^{\,|n-m|}.
\label{eq:biretention-one}
\end{equation}

\section{Methodology}
\label{sec:method}

\subsection{DRetHTR Architecture}
\label{sec:architecture}
Our proposed DRetHTR architecture builds upon the Retentive Network with a decoder-only architecture. A key adaptation involves fusing image and text embeddings within the decoder. For image–text fusion, only the parallel form is needed even during inference, so we reintroduce softmax to provide additional nonlinearity and effectively capture relationships among image tokens and between image and text tokens. For inter-text dependencies, we incorporate a temporal prior with a layer-wise scaled gamma value instead of softmax, enabling linearizability during inference.

\begin{figure*}[!t]
  \centering
  \begin{subfigure}[t]{0.58\textwidth}
    \centering
    \includegraphics[width=\linewidth]{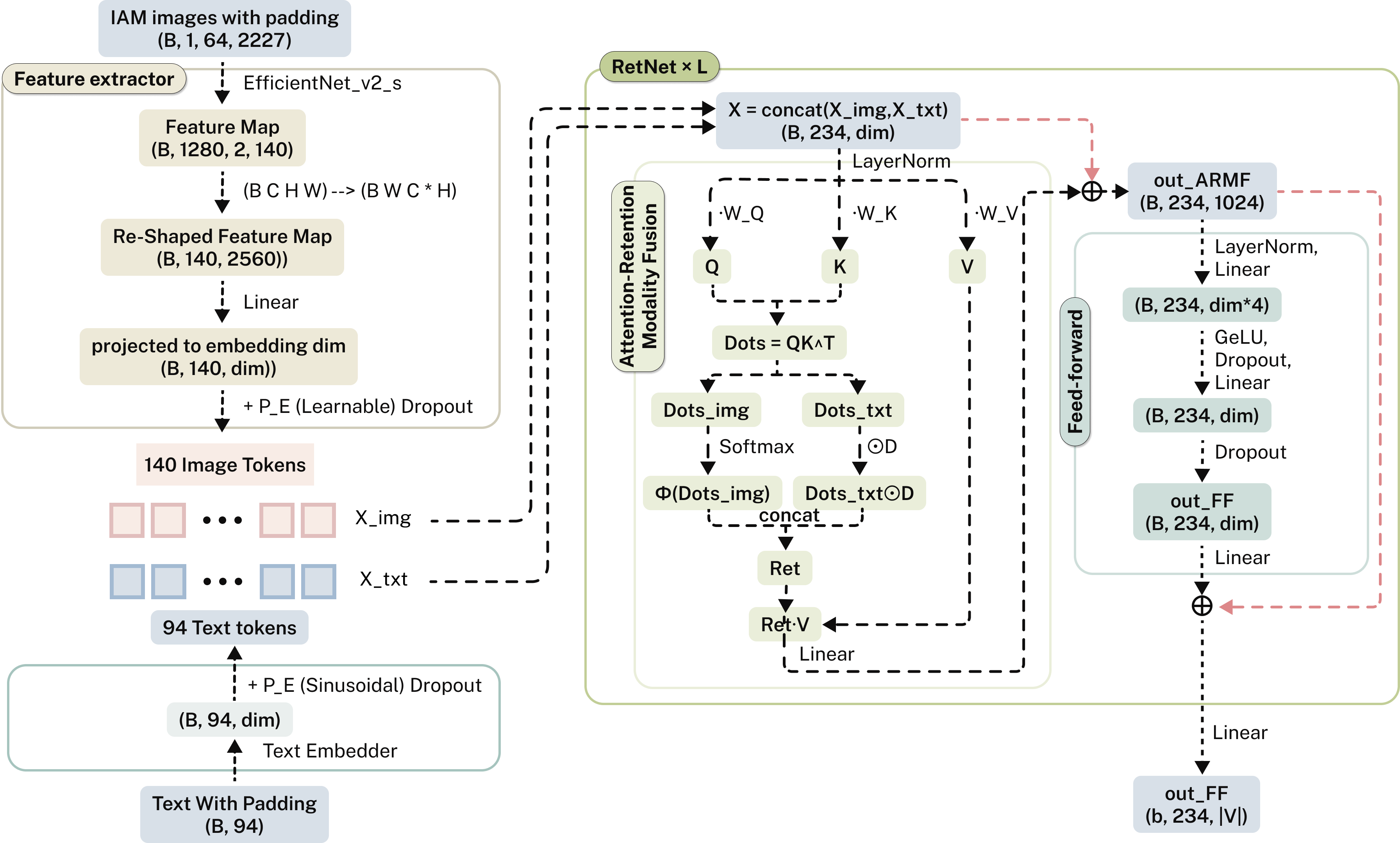}
    \caption{DRetHTR training process.}
    \label{fig:drethtr-train}
  \end{subfigure}\hfill
  \begin{subfigure}[t]{0.38\textwidth}
    \centering
    \includegraphics[width=\linewidth]{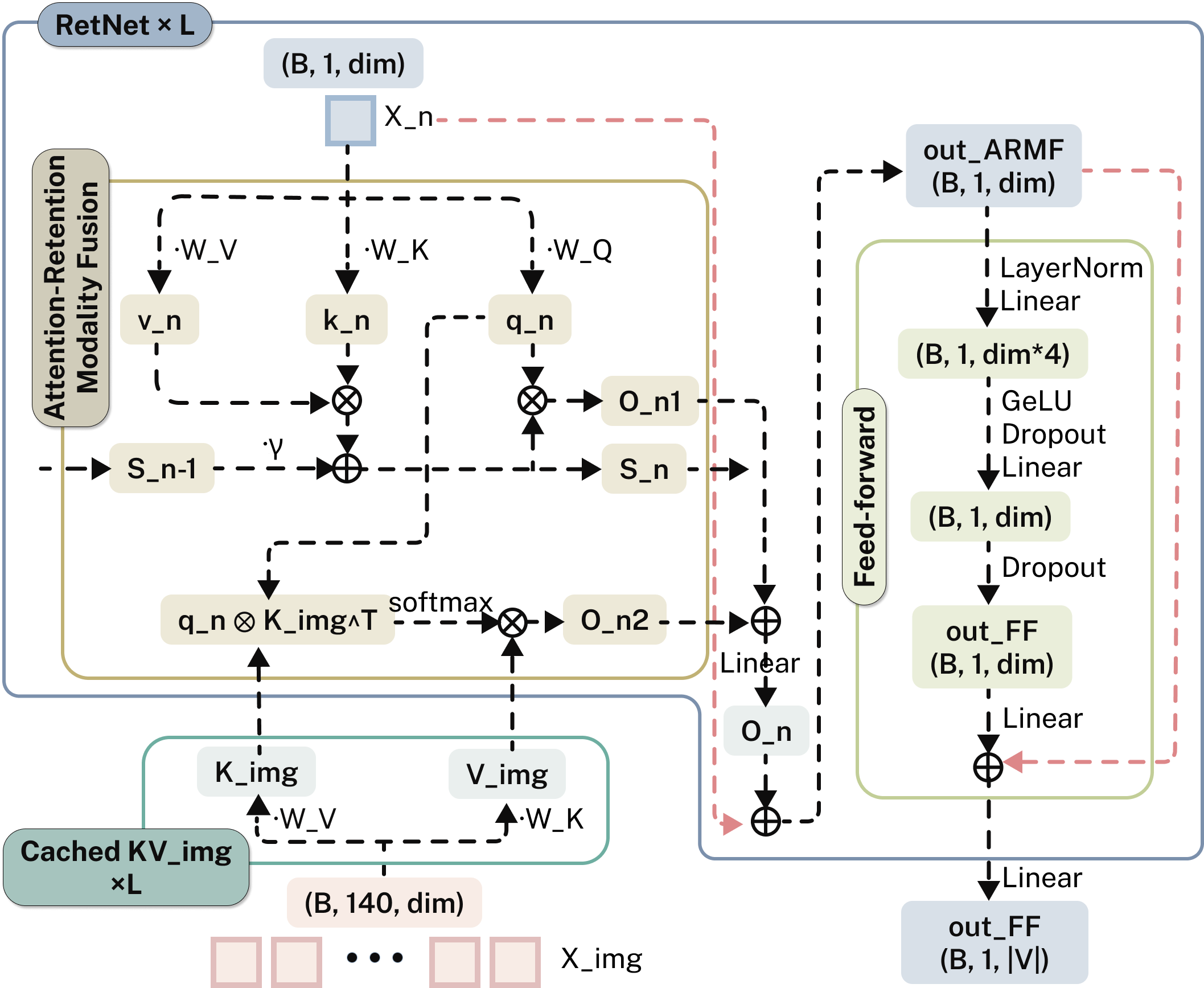}
    \caption{DRetHTR inference process.}
    \label{fig:drethtr-infer}
  \end{subfigure}

  \caption{Illustration of the DRetHTR training and inference processes: \textbf{(a)} training; \textbf{(b)} inference. ARMF mixes softmax (image) + retention (text) without breaking recurrence.}
  \label{fig:drethtr-train-infer}
\end{figure*}

\subsubsection{Image Embedding Module}
Instead of the default patch embedding, we use EfficientNetV2~\cite{Tan2021EfficientNetV2SM} as the image embedder to leverage its feature extraction capability. The ImageNet-1K pre-trained weights~\cite{deng2009imagenet} are loaded, and the classifier part is removed so that only the feature map is extracted. During training, dropout with a probability of 0.3 is applied after each
activation function across all layers. The inclusion of dropout helps the model's generalization to adapt to the handwriting domain by alleviating overfitting to the features initially learned from ImageNet. Furthermore, to obtain more spatial information for sequence processing, the stride along width of the first convolutional layer is modified from 2 to 1.
Once we obtain the 3-dimensional visual feature \({F}\) with a shape corresponding to \((c_\text{visual}, h_\text{visual}, w_\text{visual})\), where \( c_\text{visual} \) is the number of channels, and \( h_\text{visual}, w_\text{visual} \) represent the height and width of the feature map, it is reshaped into \((w_\text{visual}, c_\text{visual} \times h_\text{visual})\). Through a fully connected layer, it is then transformed into embedding \({E}\) of shape \((w_\text{visual}, d_\text{model})\), where \( d_\text{model} \) is a predefined input dimension of the decoder. Finally, the visual feature embedding is added with a learnable positional encoding, initialized with random weights. This visual feature embedding \({E}\) is considered as a sequence of \( w_\text{visual} \) image tokens, each of dimension \( d_\text{model}\).

\subsubsection{Text Embedding Module}
The transcriptions of handwriting images are tokenized at character level. The character map consists of \( V = |\mathcal{C}| \) unique characters in the dataset and the special tokens \([\text{SOS}]\), \([\text{EOS}]\), and \([\text{PAD}]\). According to the maximum length of the transcriptions in the dataset, shorter text transcriptions are padded. Finally, the input text embeddings are represented as \( {T} \) of shape \( (\max(l_{\text{text}}), d_{\text{model}}) \), where \( \max(l_{\text{text}}) \) is the fixed sequence length after padding, and \( d_{\text{model}} \) is the model dimension. Sinusoidal positional encoding is added to the embeddings to provide the model with information about the order of tokens within the sequence.

\subsubsection{DRetHTR Decoder}
After attaining the image and text tokens from their respective embedders, they are concatenated into a single sequence and passed to the RetNet decoder. The decoder consists of  \(L\) identical layers stacked on each other. Unlike encoder-decoder architectures, this design omits cross-attention. Instead, modality fusion is achieved solely within the retention layer, adapted as Attention-Retention Modality Fusion (ARMF), a mixture of multi-head attention and multi-scale retention. The architecture comprises two sub-layers: ARMF and Position-wise Feed-Forward (PFF). Both sub-layers are surrounded by residual connections, with layer normalization applied afterward. To accommodate residual connections, all sub-layers and embedding layers are designed to output the same dimension of $d_\text{model}$. The PFF consists of two linear layers, GELU for smooth non-linearity and a dropout layer to improve generalization.

\subsubsection{Attention-Retention Modality Fusion (ARMF)}
\label{sec:armf}
The limitation of full self-attention over concatenated image–text tokens is a growing KV cache and incurring \(O(N)\) per-token decoding. In contrast, applying full retention to all tokens yields \(O(1)\) decoding but weakens accuracy by limiting flexible image–text coupling. Therefore, we introduce \emph{ARMF}, which resolves this by using retention only for the autoregressive text stream—where recurrence is needed—while keeping softmax attention for image–image and image–text interactions. Image keys and values (of length $N_I$) are computed once in parallel and text stream is processed recurrently, so cross-modal softmax does not introduce a KV cache that grows with $N_T$, preserving RetNet’s \(O(1)\) decoding with respect to generated text length.
\par\smallskip

\noindent\textit{The Parallel Representation of ARMF.} As shown in \cref{fig:drethtr-train}, the ARMF layer is:
\begin{equation}
\begin{gathered}
X=\begin{bmatrix}X_{\text{img}}\\ X_{\text{text}}\end{bmatrix}\in\mathbb{R}^{N\times d_{\text{model}}},\quad N=N_I+N_T,\\
Q=XW_Q,\quad K=XW_K,\quad V=XW_V,\\
\text{Dots}=\frac{QK^\top}{\sqrt{d_k}}
=\Big[\;\text{Dots}_{\text{img}}\in\mathbb{R}^{N\times N_I}\;\; \text{Dots}_{\text{text}}\in\mathbb{R}^{N\times N_T}\;\Big],\\
D_{ij}=
\begin{cases}
0, & 1 \le i \le N_I, \\[2pt]
\gamma^{i-N_I-j}, & N_I < i \le N,\quad 1 \le j \le i-N_I, \\[2pt]
0, & N_I < i \le N,\quad i-N_I < j \le N_T.
\end{cases}\\
\text{Ret}=\Big[\;\text{softmax}(\text{Dots}_{\text{img}})\;\; \text{Dots}_{\text{text}}\odot D\;\Big]\\
\text{ARMF}(X)=\text{Ret}\,V\in\mathbb{R}^{N\times d_{\text{model}}}.
\end{gathered}
\end{equation}

Where \(X\) is the given input sequence. \(N_I\) and \(N_T\) are the numbers of image and text tokens, respectively. \(W_Q, W_K, W_V \in \mathbb{R}^{d_{\text {model }} \times d_{\text {model }}}\) are learnable weights. \(D \in \mathbb{R}^{N\ \times N_T}\) is a dependency mask that prevents image queries from attending to text keys, while, for text queries, it applies causal, exponentially decayed dependence over text keys.

Here, the reintroduced softmax does not prevent the recurrent representation, since computation for all image tokens in the inference phase is inherently parallelized. Softmax operates row-wise, so when a new text token \( q_n \) attends the full image length \( K_{\text{img}} \) and \( V_{\text{img}} \), attention with softmax can be computed solely for \( q_n \), independent of previous tokens $q_1, q_2, \dots, q_{n-1}$. Therefore, we can seamlessly expand the recurrent representation for concatenated image-text input, maintaining compatibility without any collapse.

\par\smallskip
\noindent\textit{The Recurrent Representation of ARMF} As shown in Figure~\ref{fig:drethtr-infer}, ARMF can also be computed recurrently during inference. Unlike conventional retention mechanisms, \(q_n\) must pay attention not only to the text tokens but also to the relevant image tokens. To achieve this, \(K_{\text{img}}\) and \(V_{\text{img}}\) are cached for all $L$ layers at once at the beginning of decoding, as defined below:
\begin{equation}
\begin{gathered}
\text{For } l \in [1,2, \dots, L]: \\
K_{\mathrm{img}}^{(l)} = X_{\mathrm{img}}^{(l)} \cdot W_K^{(l)}, \quad
V_{\mathrm{img}}^{(l)} = X_{\mathrm{img}}^{(l)} \cdot W_V^{(l)}
\end{gathered}
\end{equation}
Using the cached \(K_{\text{img}}\) and \(V_{\text{img}}\), text tokens can be generated autoregressively in the same manner as conventional retention mechanisms. This process is defined as follows:
\begin{equation}
\begin{gathered}
{S}_n = \gamma \cdot {S}_{n-1} + {k}_n^\top \cdot {v}_n \\
\text{o}_{\text{text}, n} = {q}_n \cdot {S}_n \in \mathbb{R}^{1 \times d_{\text{model}}} \\
\text{o}_{\text{img}, n} = \operatorname{softmax}\left(\frac{{q}_n \cdot {K}_{\text{img}}^\top}{\sqrt{d_k}}\right) \cdot {V}_{\mathrm{img}} \\
\text{ARMF}(x_n) = \text{o}_{\text{text}, n} + \text{o}_{\text{img}, n}, \quad n=1, \dots, |x|
\end{gathered}
\end{equation}
Thus, the per-step decoding cost is $O(N_I)$ and does not grow with the generated text length $N_T$; in our setting where $N_I$ is fixed, this is effectively $O(1)$ per step. Consequently, the total decoding complexity is reduced from $O(N_T^2 + N_I N_T)$ to $O(N_I N_T)$, which is linear in $N_T$.

\subsection{Multi-Scale ARMF (MARMF) with Layer-wise Gamma Scaling}
\label{sec:marmf}
Instead of using a uniform \(\gamma\) across all layers as in the original Multi-scale Retention, our ARMF employs a layer-wise gamma scaling strategy. We define this process as follows:
\begin{equation}
\label{eq:marmf}
\begin{gathered}
\gamma_{l,h} = 1 - \gamma_{\text{subtractor}}\!\left(1 - \frac{l-1}{L-1}\right) - \epsilon_h, \\ \epsilon_h = 2^{-5-4(h-1)/(H-1)} \\
\operatorname{head}_h = \operatorname{ARMF}\left(X, \gamma_{l,h}\right) \\
Y = \operatorname{Concat}\left(\operatorname{head}_1, \ldots, \operatorname{head}_H\right) \\
\operatorname{MARMF}(X) = Y W_O
\end{gathered}
\end{equation}

where $l=1,\ldots,L$ and $h=1,\ldots,H$ index layers and heads. As in RetNet's multi-scale retention, $\log\epsilon_h$ is an arithmetic progression, i.e., $\epsilon_h$ decreases geometrically from $\epsilon_1=1/32$ to $\epsilon_H=1/512$. Thus, $\gamma$ increases from smaller values in lower layers to larger values in upper layers, so lower layers emphasize local details while deeper layers capture broader context. RetNet’s original block includes GroupNorm and a Swish-style gating mechanism; gating has been argued to be beneficial in attention-style models by adding nonlinearity and sparsity and mitigating attention-sink behavior \cite{qiu2025gatedattention}. In DRetHTR, we omit these components to keep the architecture minimal and stable for line-level HTR, where the primary requirement is robust image--text alignment rather than maximizing long-range linguistic expressiveness.

\paragraph*{Layer-wise gamma scaling design rationale}
This mechanism provides explicit sequential priors by exponentially decaying the influence of earlier tokens, thereby eliminating the need for softmax-based attention between text tokens. In RetNet, each retention head employs a decay factor $\gamma$ that controls how much historical context is retained. By assigning smaller $\gamma$ values to early decoder layers and progressively increasing them in deeper layers, we mimic the \emph{inductive bias} observed in standard Transformers---where shallow layers capture local dependencies, and deeper layers attend to broader context. Crucially, this multi-scale retention allows different layers to specialise in \emph{diverse-range} dependencies (local, mid-range, and global) without incurring the quadratic cost of standard attention. The layer-wise decay acts as a structural prior for sequence modeling, compensating for the loss of flexible attention weights with a principled and efficient mechanism.

\paragraph*{Softmax-free token interactions}
Practically, softmax attention is retained only for interactions involving image tokens, so that visual features are aligned and injected into the decoding context, while all text-to-text dependencies use softmax-free recurrent retention, enabling linear-time decoding with reduced memory overhead.

\section{Evaluation}
\label{sec:experiments}

\subsection{IAM Handwriting Database}
\label{sec:dataset-iam}
We conduct experiments on the line-level IAM dataset~\cite{Marti2002IAM} using the standard Aachen split (IAM-A).
Image sizes range from \(100 \times 45\) to \(2270 \times 152\), with widths typically 4--15$\times$ larger than heights.
Transcription lengths range from 4 to 93 characters, with 79 unique characters in total.
\subsection{Synthetic Dataset for Pre-training}
\label{sec:dataset-synth}

Our model was pre-trained with 17 million image-text pairs, providing visual diversity for general HTR tasks.
To achieve this, we extracted millions of lines from the CC100 dataset~\cite{cc100}, a corpus of 2 billion English text chunks.
We randomly truncated these lines to fit the distribution of lengths of the IAM-A train dataset, ranging from a minimum length of 4 to a maximum length of 93 characters.
Any emojis or characters not found in the dataset were replaced with an empty string for compatibility.
After preparing the text,  we produced the corresponding synthetic images using 11,954 different handwriting fonts from 1001 Fonts~\cite{1001fonts}.
We assigned a randomly sampled font to each text line and rendered it as a text image in that font at a size of 100, which is close to the font size found in the IAM text lines.

\subsection{Data Preprocessing and Augmentation}

\begin{figure}[t]
    \centering
    \includegraphics[width=\linewidth]{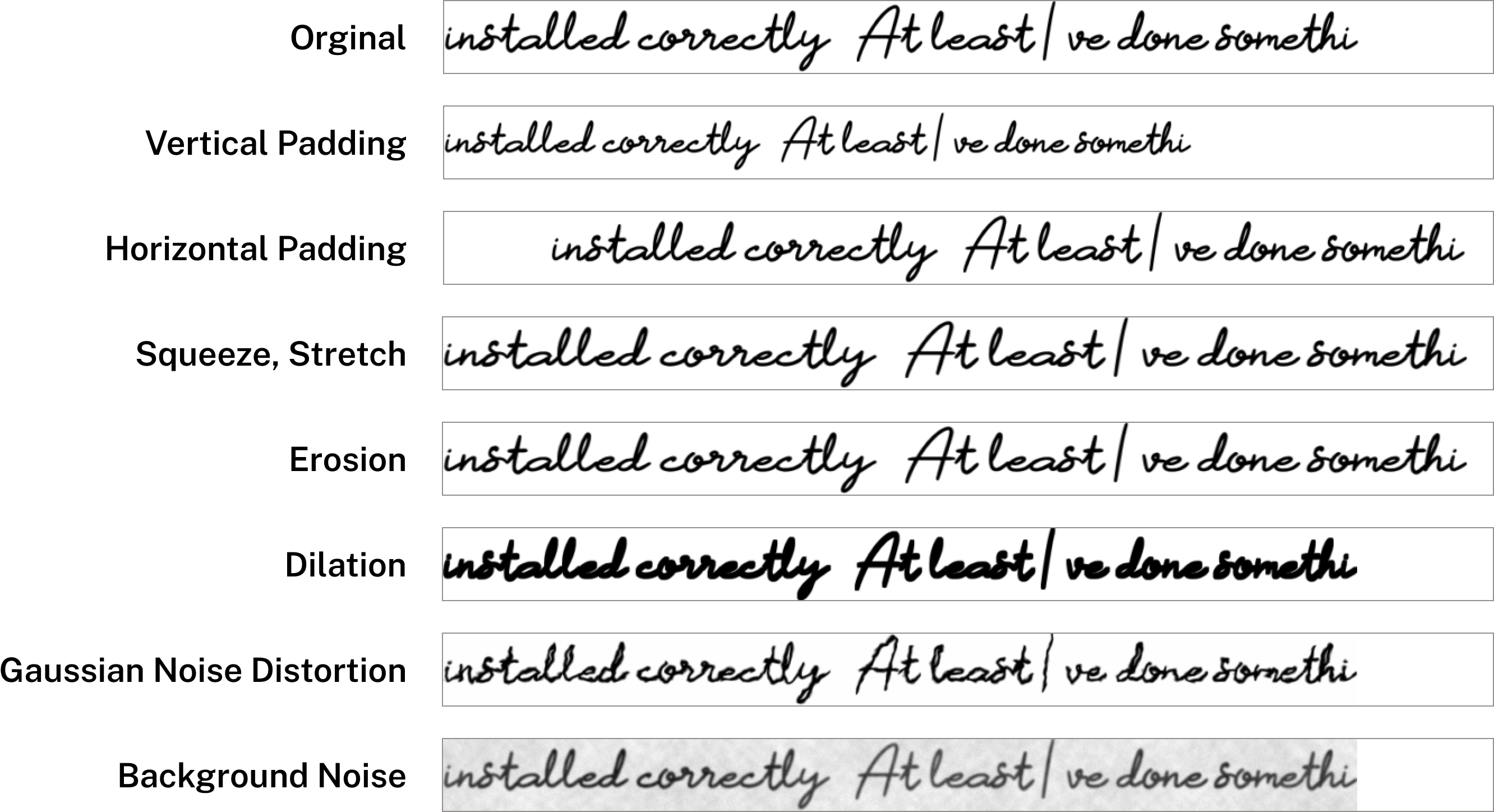}
    \caption{Examples of the six augmentations applied to a handwriting image.}
    \label{fig:augmentation}
\end{figure}

The handwriting text line images undergo a series of refined preprocessing steps. It starts by applying deslanting alone without contrast normalization. All images are resized to a constant height of 64 pixels while maintaining their original aspect ratio. However, this rescaling results in variable image widths. To perform mini-batch training, all resized images are padded with white pixels to have a uniform width of 2227 pixels. This padding is crucial, as it allows keeping the same input dimensions for the neural network. The final preprocessing step is the inversion of the pixels in the image, making the foreground (the handwritten stroke) close to 1 (white) and the background close to 0 (black). This enhances computational efficiency, increasing the prominence of features in the text while suppressing the numerical impact from the background. The dataset was augmented using six augmentation techniques, each applied with a 50\% probability. These are random padding, squeezing \& stretching, erosion, dilation, Gaussian noise distortion, background noise. Examples of each augmentation are shown in Fig.~\ref{fig:augmentation}.

\subsection{Implementation Details}
\label{sec:impl}
We use two capacity variants. DRetHTR$_{\text{WIDE}}$ has 4 decoder layers, model dimension $d_{\text{model}}=1024$, 8 heads ($d_h=128$), and PFF dimension 4096, for 73M parameters. DRetHTR$_{\text{DEEP}}$ has 12 decoder layers, $d_{\text{model}}=768$, 12 heads ($d_h=64$), and PFF dimension 3072, for 107M parameters. The names directly indicate their architectural distinction: WIDE has the larger model dimension, whereas DEEP has more decoder layers and parameters. RetHTR$_{\text{WIDE}}$, which is an encoder--decoder model, uses the same 4-layer, $d_{\text{model}}=1024$, 8-head decoder and adds a 4-layer encoder with PFF dimension 3072. All DRetHTR variants use EfficientNetV2-S as the image embedder and layer-wise scaled gamma decay for both image and text sequential priors.
The optimizer uses AdamW with $10^{-3}$ weight decay and label smoothing ($\varepsilon = 0.4$).
For the learning rate, we used a cosine annealing schedule ($10^{-4}$ to $10^{-6}$ restarting every 30 epochs).
We set dropout rates of $0.3$ for retention and \ac{pff} layers and $0.1$ for embeddings.
The pre-training on synthetic data was run on 8× NVIDIA A100 (80 GB) GPUs with a batch size of 48, while fine-tuning used a batch size of 16. Single Precision (FP32) is used throughout training and inference, following the PyTorch defaults.
To check the inference speed and memory consumption, we use a batch size of 128 with an A100 80GB GPU.
Beam size 6 is used for ablation studies and recognition comparisons because accuracy has saturated by this setting; inference-time and memory scaling are evaluated over beam sizes 1--10. Unless otherwise stated, all CER, speed, and memory results are averaged over five independent runs.

\subsection{Ablation Studies}
\label{sec:ablation}
\subsubsection{Decoder-Only vs Encoder–Decoder}
We compare both of our Decoder-Only RetNet (DRetHTR) and Encoder–Decoder RetNet (RetHTR) architectures using the WIDE 4-layer configuration. RetHTR also has a similarly configured 4-layer encoder with a 3072 hidden-size \ac{pff}. Without pre-training, RetHTR achieves slightly better results, as can be seen in Table \ref{tab:comparison}. However, DRetHTR outperforms RetHTR with pre-training on the synthetic dataset. This highlights the strength of the decoder-only approach when data-rich pre-training is used. Moreover, Table~\ref{tab:model_performance} shows that the decoder-only architecture is 3.2x faster and uses 22\,\% less memory than the RetNet version. This aligns with the expectation that removing the encoder reduces computational cost.

\begin{table}[t]
\centering
\caption{Comparison of CER and WER between DRetHTR\textsubscript{WIDE} and RetHTR\textsubscript{WIDE} on different datasets.}
\vspace{3pt}
\begingroup
\setlength{\tabcolsep}{5pt} %
\label{tab:comparison}
{\small
\begin{tabular}{@{}lcccc@{}}
\toprule
 & \multicolumn{2}{c}{\textbf{DRetHTR\textsubscript{WIDE}}} & \multicolumn{2}{c}{\textbf{RetHTR\textsubscript{WIDE}}} \\
\cmidrule(lr){2-3} \cmidrule(lr){4-5}
\textbf{Dataset} & \textbf{CER} $\downarrow$ & \textbf{WER} $\downarrow$ & \textbf{CER} $\downarrow$ & \textbf{WER} $\downarrow$ \\
\midrule
IAM         & 4.49            & 13.05           & 4.45            & 12.93 \\
IAM + Synth & \textbf{2.97}   & \phantom{1}9.13 & 3.01            & \phantom{1}9.20 \\
\bottomrule
\end{tabular}
}
\endgroup
\end{table}
\vspace{6pt}

\begin{table}[t]
\centering
\caption{Model performance comparison showcasing memory and time usage for beam-search decoding on the IAM test set (Aachen split). We refer to our Transformer baselines as TrHTR (encoder–decoder) and DTrHTR (decoder-only), which are matched reimplementations of RetHTR/DRetHTR and distinct from TrOCR/DTrOCR. Times correspond to a single pass over the full IAM test set with beam size 6 and batch size 128 on a single NVIDIA A100 80\,GB GPU. }
\vspace{3pt}
{\small
\begin{tabular*}{\linewidth}{@{\extracolsep{\fill}}lccc@{}}
\toprule
\textbf{Model} & \textbf{Params} & \textbf{\makecell{Memory\\(beam=6) $\downarrow$}} & \textbf{\makecell{Time\\(beam=6) $\downarrow$}} \\
\midrule
DRetHTR\textsubscript{WIDE} & \phantom{1}73M & \phantom{1}\textbf{9GB}  & \phantom{1}\textbf{52.8s} \\
RetHTR\textsubscript{WIDE}  & 136M           & 11.6GB                   & 170.4s \\
DTrHTR\textsubscript{WIDE}  & \phantom{1}73M & 15GB                     & \phantom{1}64.5s \\
TrHTR\textsubscript{WIDE}   & 133M           & 16GB                     & 189.4s \\
\bottomrule
\end{tabular*}
\label{tab:model_performance}
}
\end{table}

\subsubsection{Impact of different CNN feature extractors}
As shown in Table~\ref{tab:cnn_cer_comparison}, the ResNet50, used as a feature extractor of Kang~\textit{et al.}~\cite{Kang}, obtained a CER of $4.57$ with pre-training and $14.5$ without pre-training in our decoder-only architecture. The ShallowCNN, introduced by Wick~\textit{et al.}~\cite{Wick}, outperformed the ResNet50 on our DRetHTR with a CER of $4.23$ when pre-trained.
\begin{table}[t]
\centering
\caption{Comparison of CER (\%) on IAM and IAM + Synth datasets for different CNN architectures.}
\vspace{3pt}
{\small
\begin{tabular}{lcc}
\toprule
 & \multicolumn{2}{c}{\textbf{CER (\%)} $\downarrow$} \\
\cmidrule(lr){2-3}
\textbf{CNN} & \textbf{IAM} & \textbf{IAM + Synth} \\
\midrule
ResNet50 & 14.5 & 4.57 \\
ShallowCNN & 30.0 & 4.23 \\
EfficientNetV2-S & \textbf{4.49} & \textbf{2.97} \\
\bottomrule
\end{tabular}
\label{tab:cnn_cer_comparison}
}
\end{table}
However, EfficientNetV2-S excels both with and without synthetic pre-training. Trained only on IAM, it reaches a CER of 4.49\,\%, far below ResNet‑50 (14.5\,\%) and the shallow CNN (30\,\%), suggesting its Mobile Inverted Bottleneck Convolution (MBConv) + Squeeze and Excitation (SE) \cite{Tan2021EfficientNetV2SM} design captures stroke cues even in low-data regimes. With pre-training, EfficientNetV2‑S drops to 2.97\,\%, again ahead of ResNet‑50 (4.57\,\%) and the shallow CNN (4.23\,\%). In practice the backbone progressively expanding receptive fields yield multi-scale, character-specific features that are crisper than those from ResNet’s early, heavy down-sampling, making EfficientNetV2 a strong drop-in visual backbone for decoder-only HTR.

\begin{table*}[t]
\centering
\caption{Ablation of $\gamma$ scheduling on IAM (Aachen split) with an EfficientNetV2-S image encoder. All CER values are measured on the test set of Aachen split. Layer-wise $\gamma$ scaling (last row) matches the Transformer’s CER (4.49\,\%) while retaining RetNet’s \emph{softmax-free} token interactions, enabling \emph{linear-time, linear-memory} decoding. See Fig.~\ref{fig:attn-visual} for the local$\rightarrow$global progression that motivates the schedule.}
\vspace{3pt}
\fontsize{9pt}{10pt}\selectfont %

\setlength{\tabcolsep}{3.5pt} %
\renewcommand{\arraystretch}{1.3} %

\begin{tabular}{@{}l l c@{}}
\toprule
\textbf{Strategy} & \textbf{Gamma Value} & \textbf{CER} $\downarrow$ \\
\midrule
\textbf{Original} &
$\gamma_h = 1 - \epsilon_h$ & 4.65 \\
\textbf{+ Data dependent gating} &
$\gamma=\operatorname{sigmoid}(X W_\gamma)^{1 / \tau}$ & 4.66 \\
\textbf{+ Small Gamma Only} &
$\gamma_h = 1 - \gamma_{\text{subtractor}} - \epsilon_h$ & 4.60 \\
\textbf{+ Head-wise Gamma Scaling} &
$\gamma_h=\left(1-\gamma_{\text {subtractor }}-\frac{1}{32}\right)+\frac{h-1}{H-1} \gamma_{\text {subtractor }}$ & 4.51 \\
\textbf{+ Layer-wise Gamma Scaling} &
$\gamma_{l,h} = 1 - \gamma_{\text{subtractor}}\!\left(1 - \frac{l-1}{L-1}\right) - \epsilon_h$ & \textbf{4.49} \\
\textbf{Transformer with Softmax} & -- & \textbf{4.49} \\
\bottomrule
\end{tabular}
\label{tab:gamma_cer_comparison}
\end{table*}

\subsubsection{Sequential Priors for Text}
\label{sec:seq_priors_text}
In RetNet, the decay factor $\gamma \in (0,1)$ imposes a sequential prior that removes the need for softmax. The original recommendation sets $\gamma$ close to $1$ with mild head-wise variation, favoring long-range memorization. gRet further introduces a data-dependent $\gamma$ via a temperature term $\tau$, which also tends to keep $\gamma$ near $1$. While this mechanism can learn how dependency strength should decay with distance, it still treats dependencies largely at a global scale, even when only local evidence is needed, and this global bias can hinder performance on text lines where locality is often sufficient.

\begin{figure}[t]
  \centering

  \begin{subfigure}{\linewidth}
    \centering
    \includegraphics[width=\linewidth]{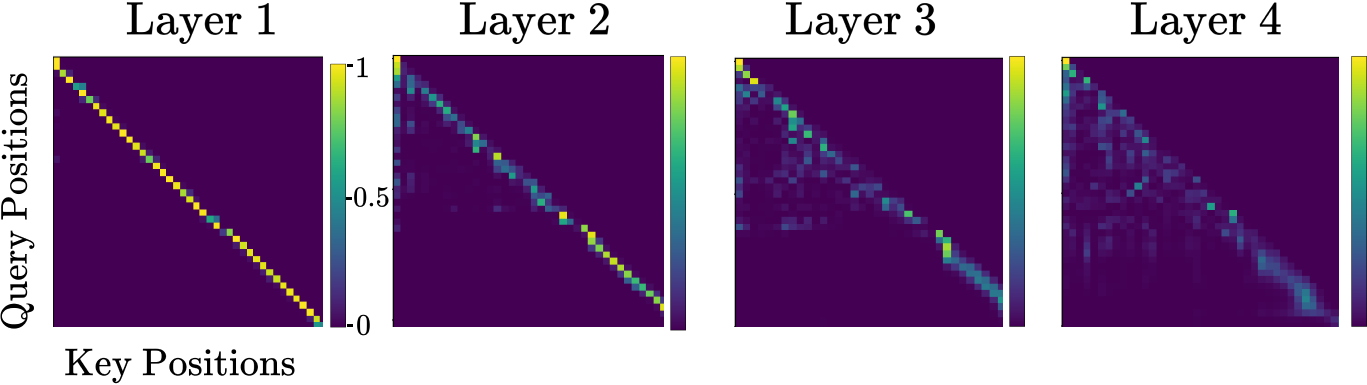}
    \caption{Self-attention matrices of DTrHTR across decoder layers: shallow layers concentrate near the diagonal (local context), while deeper layers broaden to longer-range context.}
    \label{fig:attn-visual}
  \end{subfigure}

  \vspace{4pt}

  \begin{subfigure}{\linewidth}
    \centering
    \includegraphics[width=\linewidth]{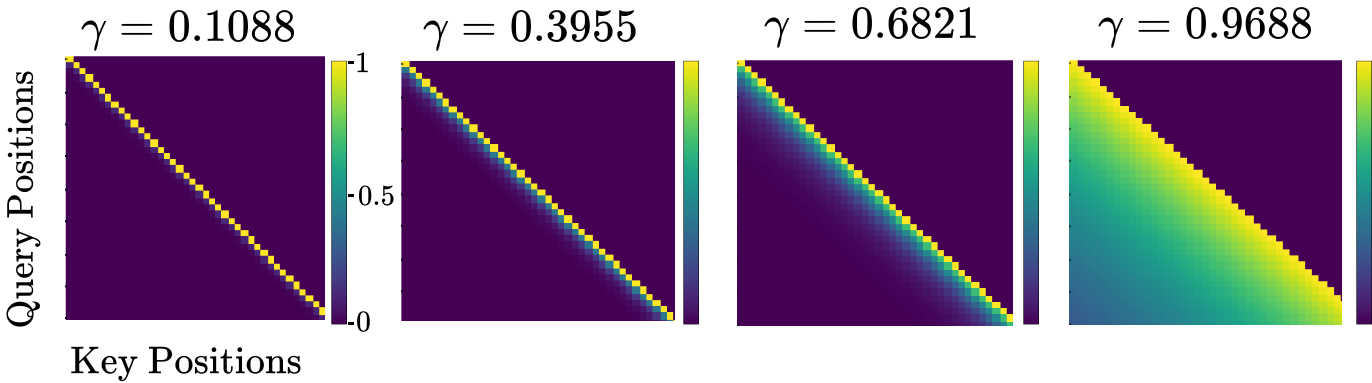}
    \caption{Retention decay matrices $D$ for the corresponding $\gamma$ settings, showing how smaller $\gamma$ emphasizes locality and larger $\gamma$ increases memory, as in the proposed layer-wise schedule.}
    \label{fig:ret-visual}
  \end{subfigure}

  \vspace{-4pt}

  \caption{Local-to-global progression and its retention mimic. \textbf{(a)} In the Transformer baseline (DTrHTR), attention naturally shifts from local to broader context with depth. \textbf{(b)} By assigning smaller $\gamma$ to early decoder layers and larger $\gamma$ to deeper layers, retention reproduces a similar progression without softmax for text–text interactions (cf. Table~\ref{tab:gamma_cer_comparison}).}
  \label{fig:trhtr-attn-ret}
\end{figure}

By contrast, a Transformer with softmax attention can flexibly emphasize either local or global dependencies as needed. In our \ac{dtrhtr}, we observe that attention often skews local rather than uniformly global. As shown in Fig.~\ref{fig:attn-visual}, the first layer concentrates on near-diagonal entries (the token and its neighbors), while deeper layers progressively broaden their receptive field. To mirror this behavior with retention, we assign smaller $\gamma$ in early decoder layers and increase it in deeper layers. The corresponding $D$-matrix visualizations for different $\gamma$ values are provided in Fig.~\ref{fig:ret-visual}.

Several $\gamma$ schedules and their CERs are summarized in Table~\ref{tab:gamma_cer_comparison}, where $\epsilon_h$ is the head-wise offset in Eq.~(\ref{eq:marmf}). We evaluate DRetHTR$_{\text{WIDE}}$ with an EfficientNetV2-S image embedder on the IAM Aachen split without the synthetic set, using $\gamma_{\text{subtractor}}=0.86$ tuned on IAM-val. Simply lowering all heads’ $\gamma$ to emphasize locality already improves over the original schedule. Introducing \emph{layer-wise} variation yields a further gain, matching the \emph{Transformer with softmax} at \textbf{4.49\,\% CER} while keeping text–text interactions \emph{softmax-free}.

\subsubsection{Sequential priors for Image}
As introduced in~\cite{rmt}, adding spatial priors in Vision Transformers can provide a boost in performance.
The default implementation here applies softmax attention between image tokens with no spatial priors.
For the CNN image embedder, the feature map is compressed into a single dimension.
Thus, a simple Euclidean distance-based prior is added bidirectionally.
Notably, this adjustment yielded better performance with a substantially faster convergence speed.

\begin{table}[t]
  \centering
  \caption{Effect of Sequential Priors on image tokens. RetNet’s temporal decay is applied bidirectionally, and layer-wise gamma scaling is added to it.}
\vspace{3pt}
    {\small
  \begin{tabular}{@{}lc@{}}
    \toprule
    \textbf{Sequential Priors for Image} & \textbf{CER} $\downarrow$ \\
    \midrule
    Priors only on text & 4.49 \\
    \hspace{0.25em}+ RetNet’s temporal decay on image & 4.41 \\
    \hspace{0.5em} + Layer-wise gamma scaling on image & 4.35 \\
    \bottomrule
  \end{tabular}
  \label{tab:sequential_priors}
  }
\end{table}
\vspace{6pt}
\paragraph*{Tokenization}
Table~\ref{tab:tokenization_comparison} compares Character-Level and Byte Pair Encoding (BPE) tokenization methods for both zero-shot and fine-tuned scenarios.
BPE yields a better zero-shot performance for CER and WER than Character-Level tokenization. However, when fine-tuning with the IAM dataset, Character-level tokenization yields a better CER, while Byte Pair Encoding results in a better WER.

\begin{table}[t]
\centering
\caption{Comparison of Character Error Rate (CER) and Word Error Rate (WER) of DRetHTR\textsubscript{DEEP} for different tokenization methods in both zero-shot and fine-tuned scenarios on the IAM test set (Aachen split). BPE uses a vocabulary size 500. BPE improves zero-shot via shorter sequences; character-level wins after fine-tuning due to precise character mapping in HTR.}
\vspace{3pt}
{\small
\begin{tabular}{lcc}
\toprule
\textbf{Tokenization} & \textbf{CER (\%) $\downarrow$} & \textbf{WER (\%) $\downarrow$} \\
\midrule
\multicolumn{3}{c}{\textit{Zero-shot Performance}} \\
\midrule
Character-Level & 11.72 & 42.09 \\
BPE $|\mathcal{V}|=500$ & \phantom{1}\textbf{9.12} & \textbf{19.87} \\
\midrule
\multicolumn{3}{c}{\textit{Fine-tuned on IAM-A}} \\
\midrule
Character-Level & \textbf{2.26} & 6.55 \\
BPE $|\mathcal{V}|=500$ & 2.36 & \textbf{4.81} \\
\bottomrule
\end{tabular}
\label{tab:tokenization_comparison}
}
\end{table}

\subsection{Qualitative error analysis on IAM}
\label{sec:error_analysis}
\begin{figure}[t]
  \centering
  \includegraphics[width=\linewidth]{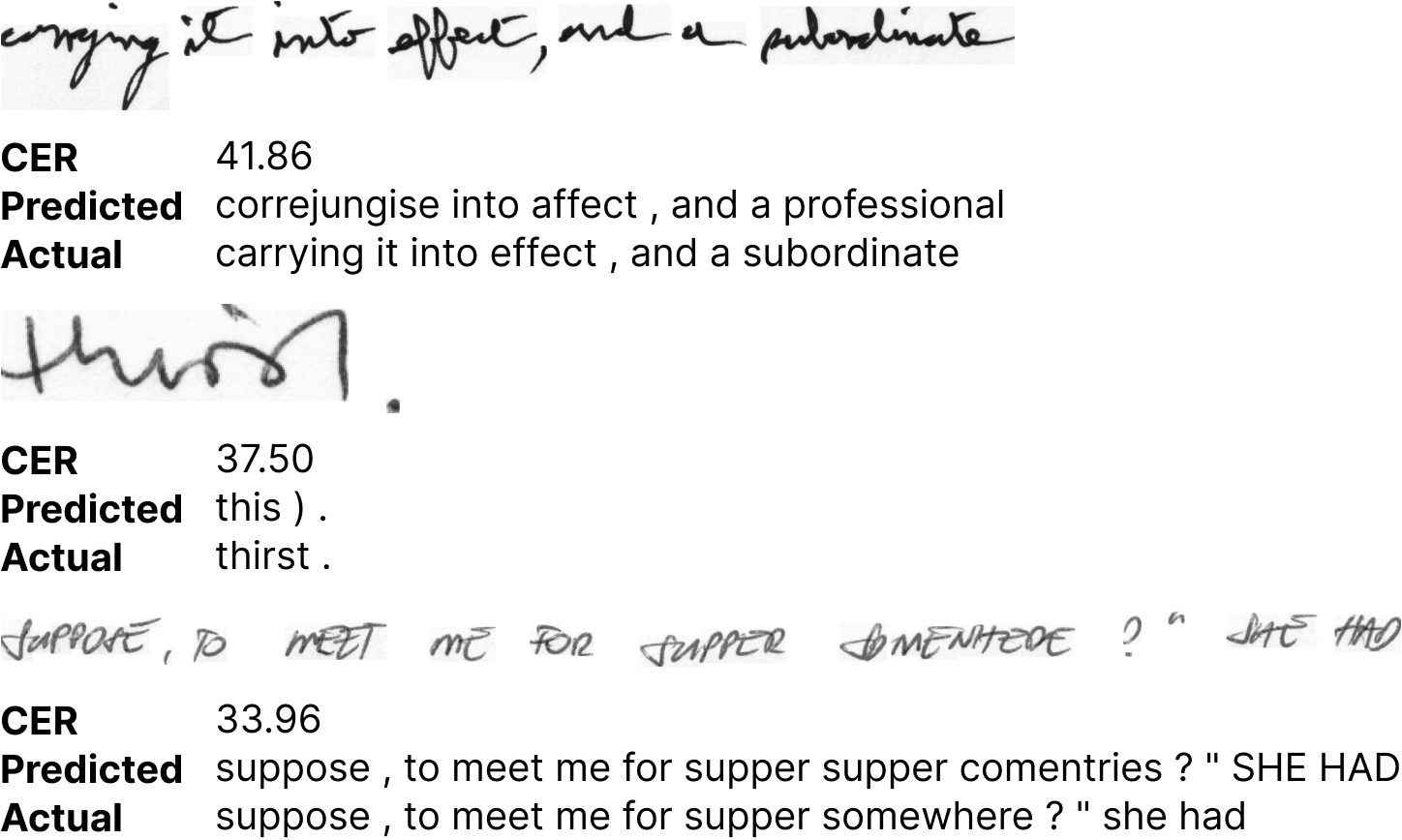}
  \caption{The three hardest IAM lines for DRetHTR (highest CER), excluding faulty labels.}
  \label{fig:error_top_lines}
\end{figure}

Figure~\ref{fig:error_top_lines} plots the three IAM lines with the highest CER after excluding samples whose ground truth is faulty. Lines 1–2 are heavily cursive, with strokes so entangled that even humans struggle, so large errors are unsurprising. Line 3 is more telling: the image is entirely uppercase, yet DRetHTR writes the sentence mostly in lowercase, capitalizing only the last two words. This shows the decoder isn’t just copying glyph shapes; it invokes RetNet's language prior to produce the more natural casing for English prose. However, this behavior departs from the visual evidence and is undesirable when case-faithful transcription is required, as discussed in Section~\ref{sec:limit}.

\subsection{Generalization beyond IAM}
\label{sec:generalization_beyond_IAM}
To assess whether DRetHTR generalizes beyond IAM, we evaluate it on nine additional line-level handwriting-recognition benchmarks and nine scene-text benchmarks. For the handwriting datasets, DRetHTR is initialized from the same English synthetic pre-training checkpoint and fine-tuned separately using the preprocessing, augmentation, and decoding settings described in \cref{sec:impl}. No external language model or lexicon is used.

\paragraph*{Evaluation across handwriting benchmarks}
\begin{table}[t]
\centering
\caption{Comparison of CER (\%) across handwriting benchmarks under the corresponding dataset protocols. The LAM leave-decade-out result is averaged across six held-out decades.}
\vspace{3pt}
\small
\footnotesize\setlength{\tabcolsep}{3pt}
\begin{tabularx}{\linewidth}{@{}>{\raggedright\arraybackslash}Xcl@{}}
\toprule
\textbf{Dataset / protocol} & \textbf{\makecell{DRetHTR\\DEEP}} & \textbf{Best published} \\
\midrule
IAM (Aachen)~\cite{Marti2002IAM} & \textbf{2.26} & 2.38 (DTrOCR~\cite{DTrOCR}) \\
RIMES~\cite{RIMES} & \textbf{1.81} & 1.91 (VAN~\cite{VAN}) \\
READ-2016~\cite{READ2016} & 4.21 & \textbf{3.59} (VAN~\cite{VAN}) \\
\midrule
Washington~\cite{Fischer2020IAMHistDB} & \textbf{1.96} & 3.86 (VQ-HTR~\cite{Davoudi2023VQHTR}) \\
Parzival~\cite{Fischer2020IAMHistDB} & \textbf{1.06} & 4.64 (VQ-HTR~\cite{Davoudi2023VQHTR}) \\
Saint Gall~\cite{Fischer2020IAMHistDB} & \textbf{2.38} & 6.79 (VQ-HTR~\cite{Davoudi2023VQHTR}) \\
\midrule
LAM, leave-decade-out average~\cite{Cascianelli2022LAM} & \textbf{2.52} & 2.9 (OrigamiNet$_{18}$) \\
ScribbleLens~\cite{Dolfing2020ScribbleLens} & \textbf{4.57} & 6.47 (CNN--BLSTM) \\
BullingerDB, frequent writers~\cite{Peer2026BullingerDB} & \textbf{7.09} & 8.2 (TrOCR) \\
BullingerDB, non-frequent writers~\cite{Peer2026BullingerDB} & \textbf{7.94} & 9.6 (TrOCR) \\
\bottomrule
\end{tabularx}
\label{tab:extended_historical_sota}
\end{table}

These benchmarks test complementary forms of generalization. IAM provides the main large-writer benchmark; RIMES tests cross-language and document variability; and READ-2016 assesses robustness to archival degradation and historical script. Washington tests low-data historical recognition, Parzival tests medieval language and script, and Saint Gall tests very early Latin manuscript text with short words. LAM's leave-decade-out protocol explicitly evaluates temporal handwriting evolution within one writer across six decades~\cite{Cascianelli2022LAM}; ScribbleLens adds broad historical writer variability across more than 80 writers~\cite{Dolfing2020ScribbleLens}; and BullingerDB tests large-scale multilingual recognition for both frequent and rare or unseen writers~\cite{Peer2026BullingerDB}. DRetHTR$_{\text{DEEP}}$ achieves the best listed CER on IAM, RIMES, Washington, Parzival, Saint Gall, LAM, ScribbleLens, and both BullingerDB subsets, while remaining competitive on READ-2016.
\paragraph*{Evaluation across scene-text benchmarks}
To assess transferability beyond handwriting, we fine-tune DRetHTR on the standard Synth protocol using MJSynth~\cite{Jaderberg2014MJSynth} and SynthText~\cite{Gupta2016SynthText}. We evaluate word-recognition accuracy on eight test sets from IIIT5K~\cite{mishra2012iiit5k}, SVT~\cite{wang2011svt}, ICDAR 2013/2015~\cite{karatzas2013icdar,karatzas2015icdar}, SVTP~\cite{phan2013svtp}, and CUTE~\cite{risnumawan2014cute80}. Table~\ref{tab:scene_text} reports the epoch-20 fine-tuned checkpoint decoded with beam search ($B=6$). It exceeds TrOCR$_{\text{LARGE}}$ on IIIT5K; performance on the more irregular IC15, SVTP, and CUTE sets remains challenging. Baseline values are taken from the original publications, which differ from ours in pre-training corpora, initialization (TrOCR and DTrOCR build on large pre-trained vision and language models), visual backbones, and model size. As for the IAM comparison in Section~\ref{sec:efficiency}, these cross-paper results position DRetHTR relative to published systems rather than constituting a controlled comparison of decoding architectures.

\begin{table*}[t]
\centering
\caption{Word-recognition accuracy (\%) on standard cropped scene-text benchmarks under the Synth protocol (MJSynth+SynthText). The highest accuracy in each column is shown in bold.}
\vspace{3pt}
\small
\setlength{\tabcolsep}{2pt}
\begin{tabular}{lcccccccc}
\toprule
\textbf{Method} & \textbf{IIIT5K} & \textbf{SVT} & \textbf{IC13$_{857}$} & \textbf{IC13$_{1015}$} & \textbf{IC15$_{1811}$} & \textbf{IC15$_{2077}$} & \textbf{SVTP} & \textbf{CUTE} \\
\midrule
TrOCR$_{\text{LARGE}}$~\cite{TrOCR} & 91.0 & 93.2 & 98.3 & 97.0 & 84.0 & 78.0 & 91.0 & 89.6 \\
DiffusionSTR~\cite{DiffusionSTR2023} & 97.3 & 93.6 & 97.1 & 96.4 & 86.0 & 82.2 & 89.2 & 92.5 \\
\textbf{DRetHTR$_{\text{DEEP}}$ (Ours)} & 97.1 & 93.9 & \textbf{98.9} & \textbf{98.2} & 84.5 & 81.3 & 88.6 & 91.11 \\
DTrOCR~\cite{DTrOCR} & \textbf{98.4} & \textbf{96.9} & 98.8 & 97.8 & \textbf{92.3} & \textbf{90.4} & \textbf{95.0} & \textbf{97.6} \\
\bottomrule
\end{tabular}
\label{tab:scene_text}
\end{table*}

\paragraph*{Evaluation on challenging non-Latin and multilingual text}
We further test cross-script transfer on SCUT-HCCDoc, which contains camera-captured Chinese handwriting with substantial perspective, background, orientation, and character variation~\cite{Zhang2020SCUTHCCDoc}. DRetHTR$_{\text{DEEP}}$ is initialized from the English synthetic-handwriting checkpoint and fine-tuned on the Chinese text lines. Accuracy rate (AR) penalizes insertions, whereas correct rate (CR) ignores them; both are character-level measures. Table~\ref{tab:scut_hccdoc} shows that DRetHTR improves over the CTC-branch output of SegCTC~\cite{Huang2023SegCTC}, demonstrating effective transfer to a large non-Latin character inventory.

\begin{table}[t]
\centering
\caption{Recognition performance on the SCUT-HCCDoc test set. Higher values are better.}
\vspace{3pt}
\label{tab:scut_hccdoc}
\small
\begin{tabular}{lcc}
\toprule
\textbf{Method} & \textbf{AR (\%)} $\uparrow$ & \textbf{CR (\%)} $\uparrow$ \\
\midrule
SegCTC (best published)~\cite{Huang2023SegCTC} & 92.08 & 93.40 \\
\textbf{DRetHTR$_{\text{DEEP}}$ (Ours)} & \textbf{92.60} & \textbf{93.90} \\
\bottomrule
\end{tabular}
\end{table}

For multilingual arbitrary-shaped scene text, we partition the released ICDAR2019-ArT training set 8:1:1; official test annotations are unavailable~\cite{Chng2019ArT}. DRetHTR obtains 9.88\% CER on the Latin subset and 18.79\% on the full multilingual subset. These diagnostic values are not compared with official competition results.

Overall, ten handwriting datasets and nine scene-text evaluations, spanning historical and modern documents, Latin and Chinese scripts, diverse writers, natural images, and irregular geometry, provide broad evidence that DRetHTR generalizes beyond IAM.

\subsection{Comparative Inference Time and Memory Efficiency}
\label{sec:efficiency}
The main benefits of RetNet are short inference time and low memory requirements. As seen in Table~\ref{tab:inference_speed_comparison}, in our implementation,  RetNet outperforms Transformer in every evaluated aspect, including inference time, memory consumption, and CER. With BPE, which makes the total tokens less than character-level tokenization, it achieves 1.5x faster inference. Additionally, compared to the TrOCR models\footnote{This code version is from the Hugging Face documentation: \url{https://huggingface.co/docs/transformers/model_doc/trocr}}, our approach provides better performance with fewer parameters but significantly improves the inference speed. The memory consumption is measured by the difference in peak allocated memory before and after decoding one batch. For the other settings, see \cref{sec:impl}.

\begin{table*}[t]
\centering
\caption{Comparative inference efficiency of RetNet, Transformer, and TrOCR models on the IAM test set (Aachen split). All times and memory values are measured for decoding the full IAM test set with the specified tokenization and beam size 10, using batch size 128 on a single NVIDIA A100 80\,GB GPU.}
\vspace{3pt}
\scriptsize %
\setlength{\tabcolsep}{3pt} %
\renewcommand{\arraystretch}{0.95} %

\begin{tabular}{lcccccc}
\toprule
\textbf{Model} & \textbf{Params} & \textbf{Tokenization} & \textbf{Total Tokens} & \textbf{Inference Time} $\downarrow$ & \textbf{Memory} $\downarrow$ & \textbf{CER} $\downarrow$ \\
\midrule
\makecell[l]{\textbf{DRetHTR$_{\text{DEEP}}$}} & 107M & Character Level & 442,832 & \textbf{123s} & 22.14GB & \textbf{2.26} \\
\makecell[l]{\textbf{DTrHTR$_{\text{DEEP}}$}} & 107M & Character Level & 442,832 & 233s & 36.25GB & \textbf{2.35} \\
\midrule\midrule
\makecell[l]{\textbf{DRetHTR$_{\text{WIDE}}$}} & \phantom{1}73M & Character Level & 442,832 & \phantom{1}67s & 14.5GB & 2.97 \\
\makecell[l]{\textbf{DTrHTR$_{\text{WIDE}}$}} & \phantom{1}73M & Character Level & 442,832 & 110s & 25.1GB & 2.98 \\
\makecell[l]{\textbf{DRetHTR$_{\text{WIDE}}$}} & \phantom{1}74M & BPE $|\mathcal{V}|=500$ & \phantom{1}49,574 & \phantom{1}44s & 14.9GB & 3.01 \\
\makecell[l]{\textbf{DTrHTR$_{\text{WIDE}}$}} & \phantom{1}74M & BPE $|\mathcal{V}|=500$ & \phantom{1}49,574 & \phantom{1}69s & 23.9GB & 3.01 \\
\makecell[l]{\textbf{TrOCR$_{\text{SMALL}}$}} & \phantom{1}62M & BPE(GPT-2) $|\mathcal{V}|=50,257$ & \phantom{1}31,959 & 310s & \phantom{1}9.1GB & 4.22 \\
\makecell[l]{\textbf{TrOCR$_{\text{BASE}}$}} & 334M & BPE(GPT-2) $|\mathcal{V}|=50,257$ & \phantom{1}31,959 & 1717s & 12.3GB & 3.42 \\
\makecell[l]{\textbf{TrOCR$_{\text{LARGE}}$}} & 558M & BPE(GPT-2) $|\mathcal{V}|=50,257$ & \phantom{1}31,959 & 2725s & 13.5GB & 2.89 \\
\bottomrule
\end{tabular}
\label{tab:inference_speed_comparison}
\end{table*}

\vspace{3pt}

\begin{table*}[t]
\centering
\caption{Comparison of HTR models: parameters, architecture, training data, external language model (LM) usage, and CER on the IAM test set (Aachen split).
Abbrev.: Syn = synthetic data; Int. = internal data; Tx = Transformer; EN-V2 = EfficientNetV2.
For architectures, $L$/$h$/$d$ denote the number of blocks (layers), attention heads per block, and embedding dimension (e.g., $L24,h16,d1024$ means 24 layers, 16 heads, 1024-dim embeddings).}
\vspace{3pt}
\label{tab:htr_model_comparison}

\resizebox{\textwidth}{!}{%
\begin{tabular}{lcccccc}
\toprule
\textbf{Model} & \textbf{Params} & \textbf{Encoder} & \textbf{Decoder} & \textbf{Train} & \textbf{LM} & \textbf{CER} $\downarrow$ \\
\midrule

\makecell[l]{\textbf{DAN}~\cite{DAN}}
& $\approx$2.18M
& \makecell[l]{23Conv+GRU}
& \makecell[l]{GRU, L1/d256}
& IAM & No & 6.40 \\

\makecell[l]{\textbf{Wick \textit{et al.}}~\cite{Wick}}
& $\approx$22M
& \makecell[l]{3Conv + Tx, L3/h8/d512}
& \makecell[l]{Bidi Tx, L3/h8/d512}
& IAM & No & 5.67 \\

\makecell[l]{\textbf{Michael~\textit{et al.}}~\cite{S2S}}
& $\approx$1.8M
& \makecell[l]{3Conv+3BLSTM, d256}
& \makecell[l]{Attn-LSTM, L1/d256}
& IAM & No & 4.87 \\

\makecell[l]{\textbf{HTR-VT}~\cite{htrvt}}
& 53.5M
& \makecell[l]{ResNet18 + Tx-Enc, L4/h6/d768}
& \makecell[l]{CTC}
& IAM
& No
& 4.7 \\

\makecell[l]{\textbf{Kang~\textit{et al.}}~\cite{Kang}}
& 100M
& \makecell[l]{Tx w/ResNet50, L4/h8/d1024}
& \makecell[l]{Tx, L4/h8/d1024}
& Syn+IAM & No & 4.67 \\

\makecell[l]{\textbf{VAN}~\cite{VAN}}
& 2.7M
& \makecell[l]{18 VGG-style Conv}
& \makecell[l]{1$\times$LSTM d256 + CTC}
& IAM
& No
& 4.45
\\

\makecell[l]{\textbf{Puigcerver}~\cite{CNNBiLSTMCTC}}
& $\approx$8.3M
& \makecell[l]{5Conv}
& \makecell[l]{5$\times$BiLSTM d256 + CTC}
& IAM & Yes & 4.4 \\

\makecell[l]{\textbf{Bluche and Messina}~\cite{Bluche2017}}
& 0.75M
& \makecell[l]{GCRNN, 6Conv+2BiLSTM, d128}
& \makecell[l]{CTC}
& Syn+IAM & No & 3.20 \\

\makecell[l]{\textbf{TrOCR$_{\text{LARGE}}$}~\cite{TrOCR}}
& 558M
& \makecell[l]{BEiT$_{\text{LARGE}}$, L24/h16/d1024}
& \makecell[l]{RoBERTa$_{\text{LARGE}}$, L12/h16/d1024}
& Syn+IAM & No & 2.89 \\

\makecell[l]{\textbf{Diaz~\textit{et al.}}~\cite{Diaz2021RethinkingTL}}
& 1.5M
& \makecell[l]{SelfAttn, L16/h4/d256}
& \makecell[l]{CTC}
& Int.+IAM & Yes & 2.75 \\

\makecell[l]{\textbf{DTrOCR}~\cite{DTrOCR}}
& 105M
& --
& \makecell[l]{GPT-2, L12/h12/d768}
& Syn+IAM & No & 2.38 \\

\midrule
\makecell[l]{\textbf{DRetHTR$_{\text{WIDE}}$} (ours)}
& 73M
& --
& \makecell[l]{RetNet w/EN-V2, L4/h8/d1024}
& IAM & No & 4.35 \\

\makecell[l]{\textbf{DRetHTR$_{\text{WIDE}}$} (ours)}
& 73M
& --
& \makecell[l]{RetNet w/EN-V2, L4/h8/d1024}
& Syn+IAM & No & 2.97 \\

\makecell[l]{\textbf{DRetHTR$_{\text{DEEP}}$} (ours)}
& 107M
& --
& \makecell[l]{RetNet w/EN-V2, L12/h12/d768}
& Syn+IAM & No & \textbf{2.26} \\
\bottomrule
\end{tabular}%
}
\end{table*}
\vspace{6pt}

Table~\ref{tab:htr_model_comparison} compares recognition accuracy and model capacity across published systems. Although every CER uses the IAM Aachen test split, the training resources differ: ``IAM'' denotes training only on IAM, ``Syn+IAM'' adds synthetic pre-training, and ``Int.+IAM'' includes non-public internal data; ``LM=Yes'' indicates use of an external language model. The two WIDE rows isolate the effect of synthetic pre-training within our pipeline, reducing CER from 4.35\% to 2.97\%. Among similarly sized decoder-only systems trained with synthetic data and IAM but no external LM, DRetHTR$_{\text{DEEP}}$ obtains 2.26\% CER with 107M parameters, compared with DTrOCR's 2.38\% at 105M. It also improves over the 558M-parameter TrOCR$_{\text{LARGE}}$ result reported under the same broad training-data category. Differences in synthetic corpora, optimization, and visual backbones prevent attributing cross-paper gaps solely to retention; the controlled DTrHTR comparison in Table~\ref{tab:inference_speed_comparison} is therefore the primary evidence for the decoder efficiency claim.

\subsection{Beam Search Efficiency of RetNet}
Beam search magnifies decoding-time overheads because all decoder states must be maintained and updated for beam size of candidate partial transcriptions in parallel. In a KV-cached Transformer, each beam stores a growing history of keys and values for every layer, and beam pruning requires reindexing these caches at every step. It also allocates temporary tensors when appending new keys/values, making decoding increasingly memory- and bandwidth-bound as beam size grows. In contrast, RetNet’s recurrent form maintains a fixed-size per-layer state for each beam and updates it in-place, avoiding sequence-length-dependent KV caches and reducing cache-management overhead.

From an asymptotic FLOP perspective (Table~\ref{tab:computations}), a KV-cached Transformer requires $O(nd)$ work per decoding step, whereas RetNet’s recurrent update scales as $O(d^2)$. This suggests that recurrent decoding is expected to be FLOP-favorable mainly when the generated length $n$ exceeds the model width $d$, as in long-sequence settings. However, wall-clock decoding on GPUs is not determined by FLOPs alone: cache growth, reindexing, and memory traffic can dominate runtime under beam search. Empirically, we find that RetNet remains faster and more memory efficient even on IAM, where sequences are relatively short ($n<d$), because the Transformer’s per-beam KV cache and its associated management overhead grow with both $n$ and beam size $B$, while RetNet’s recurrent state is fixed-size and beam-friendly.

\begin{figure}[t]
  \centering
  \includegraphics[width=\linewidth]{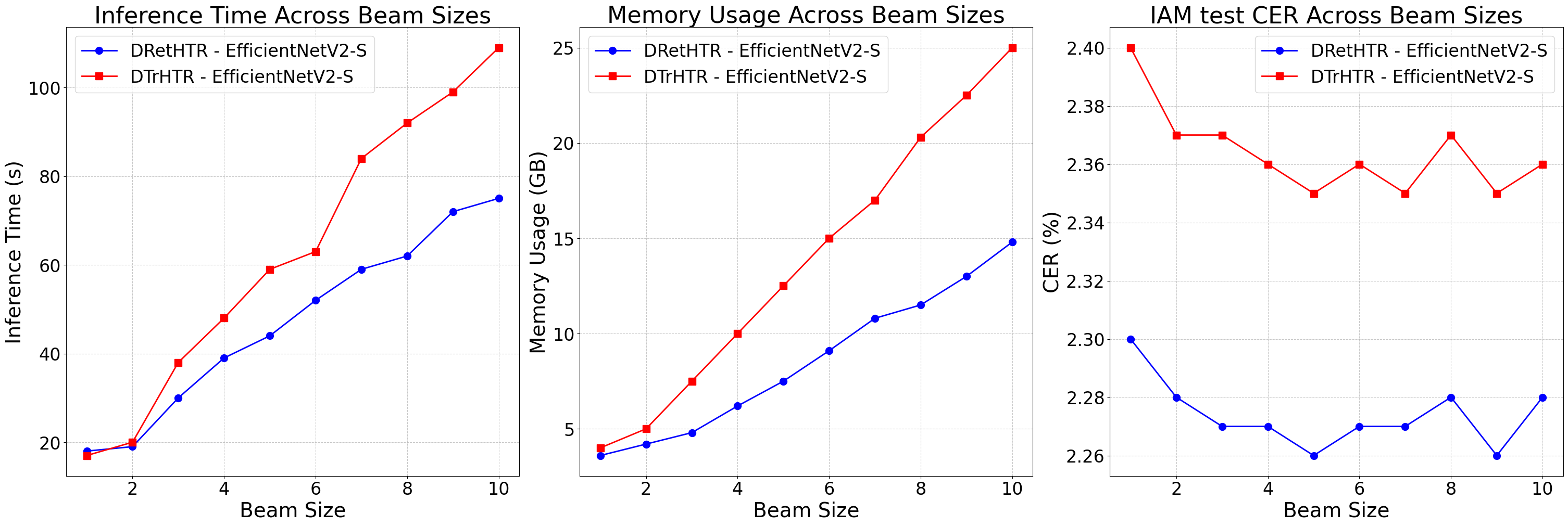}
    \caption{Beam-size scaling on the IAM Aachen test set: end-to-end inference time (left), peak memory (middle), and test CER (right) for DRetHTR vs.\ DTrHTR. DRetHTR scales more favorably with beam size because of fixed-size recurrent states, whereas KV-cached decoding incurs beam-dependent cache growth and cache reindexing after beam pruning. CER generally improves with larger beams up to $B=5$, after which it fluctuates slightly while decoding cost continues to grow with $B$.}
  \label{fig:beam1_10}
\end{figure}

Table~\ref{tab:memory_allocation} summarizes the dominant per-layer storage. For DTrHTR$_{\text{DEEP}}$ ($d{=}768$, $H{=}12$) and a representative maximum decoding length $N=94$, RetNet stores $B\,d^2/H$ elements per layer, whereas the Transformer stores $2\,B\,N\,d$ elements for KV caching. Figure~\ref{fig:beam1_10} shows that DRetHTR scales more favorably with beam size on the IAM Aachen test set. Recognition accuracy typically saturates at moderate beams (around $B{=}5$), indicating that most remaining errors stem from visual/model ambiguity rather than insufficient search. We evaluate up to $B{=}10$ for consistency with TrOCR; however, our character-level or 500-token BPE vocabulary yields a more peaked next-token distribution and a smaller branching factor than GPT-2 tokenization, so larger beams add little diversity and do not materially improve CER, while still increasing decoding cost.

\begin{table}[t]
    \centering
        \caption{Dominant per-layer state size for beam-search decoding. $d$ is the model dimension, $H$ the number of heads, $N$ the decoded length, and $B$ the beam size. Example element counts use $d{=}768$, $H{=}12$, $N{\approx}94$, and $B{=}5$ (DTrHTR$_{\text{DEEP}}$). RetNet's state is fixed-size in $N$, while KV caching scales linearly with $N$.}
\vspace{3pt}
    \footnotesize %
    \begin{tabular}{lcc}
        \toprule
        \textbf{Decoding method} & \textbf{\makecell{Memory\\per layer}} & \textbf{\makecell{Number of\\elements}} \\
        \midrule
        Recurrent (RetNet) & $B\,d^{2}/H$ & $0.25$M \\
        KV-Cached Transformer & $2\,B\,N\,d$ & $0.72$M \\
        \bottomrule
    \end{tabular}

    \label{tab:memory_allocation}
\end{table}
\vspace{10pt}

\section{Limitations}
\label{sec:limit}

A central design choice in DRetHTR is to retain softmax only for image--text fusion, while modeling text--text dependencies via softmax-free retention with sequential priors. This design yields clear efficiency gains and, in our experiments, even improves accuracy over the decoder-only Transformer baseline (DTrHTR). DTrHTR normalizes each text query over a single key set that mixes image and text tokens, so visual evidence and linguistic context compete within one probability simplex, whereas DRetHTR restricts softmax normalization for text queries to image keys. The improvement is therefore consistent with the hypothesis that joint softmax over all keys dilutes the probability mass on the most relevant image tokens, especially for visually ambiguous handwriting. However, our experiments do not include a controlled ablation isolating (i) the presence of softmax itself, (ii) the scope of its normalization (all tokens vs.\ modality-specific subsets), and (iii) its interaction with retention-style sequential priors.

In addition, our evaluation is centered on line-level benchmarks with moderate sequence lengths, and thus does not fully probe the regime $n \gg d$ where retentive decoding is expected to provide its strongest computational advantage over KV-cached attention. We also do not systematically study alignment stability under very long autoregressive decoding, where accumulated errors can lead to \emph{attention drift} and degrade visual--text correspondence.

Beyond sequence length, the line-level handwriting and cropped-word scene-text evaluations assume that each text region has already been localized. DRetHTR therefore does not address paragraph or table structure, page-level reading order, text detection, or geometric rectification of curved or irregular text; extending it to layout-rich documents or end-to-end scene-text reading would require integration with such modules.

The decoder's learned language prior also entails a trade-off. As shown in Section~\ref{sec:error_analysis}, DRetHTR may transcribe an all-uppercase line with conventional sentence casing. Such contextual correction can resolve ambiguous glyphs, but it is a drawback when strict fidelity to the source is required, e.g., diplomatic transcription of historical documents, which must preserve capitalization, abbreviations, and non-standard spellings, or case-sensitive identifiers and codes. We do not quantify this effect separately; case-balanced synthetic pre-training or explicit control of the language prior's weight could mitigate it.

Finally, layer-wise gamma scaling relies on empirically selected hyperparameters: $\gamma_{\text{subtractor}}=0.86$ was tuned on IAM-val, and $\epsilon_h$ follows RetNet. We kept this schedule fixed for all other datasets, including Chinese SCUT-HCCDoc and the scene-text benchmarks, and still obtained competitive results, suggesting that it is not specific to IAM. This is plausible because $\gamma$ controls decay over token distance rather than script-specific units: the effective horizon $1/(1-\gamma)$ ranges from about one token in the first layer to 32--512 tokens in the last, covering typical text-line lengths. Nevertheless, dataset-specific tuning might further improve accuracy, e.g., for scripts whose tokens carry more information, such as Chinese. For substantially longer sequences, e.g., paragraph- or page-level decoding beyond the largest horizon, the deeper-layer decay would likely need to increase, requiring re-tuning or a length-adaptive schedule.

\section{Future work}
\label{sec:future}

A natural next step is to factorize the normalization for text queries, e.g., by computing separate attention distributions over image and text keys (or a gated mixture of the two), and to compare these variants with and without retention-based sequential priors. This would clarify whether DRetHTR benefits primarily from restricting cross-modal competition in a single softmax, from the inductive bias of layer-wise gamma scaling, or from their interaction.

Future work should also stress-test retentive decoding on substantially longer sequences to characterize efficiency and accuracy in the $n \gg d$ regime, where recurrent decoding is expected to become decisively advantageous. Mitigating attention drift in long-form decoding is another important direction, e.g., via alignment-regularized training, lightweight alignment correction, or decoding strategies that monitor and stabilize image--text correspondence.

\section{Conclusion}
\label{sec:conclude}
We introduced DRetHTR, a linear-time, linear-memory decoder-only handwritten text recognition model based on Retentive Networks, and demonstrated that softmax-free retentive decoding achieves Transformer-level accuracy while substantially reducing inference cost. Across ten handwriting and nine scene-text benchmarks, DRetHTR attains state-of-the-art or competitive accuracy under consistent training and decoding settings, while lowering decoding latency and peak memory relative to an equally sized decoder-only Transformer baseline. These gains stem from two adaptations: ARMF retains softmax attention at the modality interface for strong image--text alignment while keeping text--text interactions causal and recurrence-compatible, and layer-wise gamma scaling restores the local-to-global inductive bias of Transformers by expanding the effective retention horizon with depth. Overall, retention is a practical alternative to KV-cached attention for HTR, combining competitive accuracy with substantially improved decoding efficiency.

\section*{\normalsize CRediT authorship contribution statement}
\textbf{Changhun Kim}: Conceptualization, Data curation, Formal analysis, Investigation, Methodology, Software, Validation, Visualization, Writing -- original draft, Writing -- review \& editing; \textbf{Martin Mayr}: Formal analysis, Methodology, Resources, Software, Validation, Writing -- review \& editing; \textbf{Thomas Gorges}: Formal analysis, Validation, Visualization, Writing -- review \& editing; \textbf{Fei Wu}: Formal analysis, Methodology, Validation, Writing -- review \& editing; \textbf{Mathias Seuret}: Formal analysis, Investigation, Validation, Writing -- review \& editing; \textbf{Andreas Maier}: Supervision, Resources; \textbf{Vincent Christlein}: Conceptualization, Formal analysis, Methodology, Project administration, Supervision, Validation, Writing -- review \& editing.

\section*{\normalsize Acknowledgements}
The authors gratefully acknowledge the scientific support and HPC resources provided by the Erlangen National High Performance Computing Center (NHR@FAU) of the Friedrich-Alexander-Universität Erlangen-Nürnberg (FAU). The hardware is partially funded by the German Research Foundation (DFG).

\section*{\normalsize Declaration of generative AI and AI-assisted technologies in the manuscript preparation process}
During the preparation of this work, the authors used ChatGPT (OpenAI) in order to improve language, readability, paragraph formulation, and LaTeX equation presentation. After using this tool, the authors reviewed and edited the content as needed and take full responsibility for the content of the published article.

\section*{\normalsize Data availability}
The benchmark datasets used in this study are publicly available from their official sources. The code used in this study, together with the access link for the synthetic pre-training data, is available at \url{https://github.com/Kimchangheon/DRetHTR}.

\section*{\normalsize Declaration of competing interests}
The authors declare that they have no known competing financial interests or personal relationships that could have appeared to influence the work reported in this paper.

\begingroup
\setlength{\itemsep}{0pt}
\setlength{\parskip}{0pt}

\let\oldthebibliography\thebibliography
\renewcommand{\thebibliography}[1]{%
  \oldthebibliography{#1}%
  \setlength{\itemsep}{0pt plus 0.3ex}%
}
\begin{small} %
\bibliographystyle{IEEEbib}
\bibliography{refs}
\end{small}
\endgroup

\clearpage
\appendix
\renewcommand{\theequation}{\thesection.\arabic{equation}}
\setcounter{equation}{0}
\appsection{Connection between RetNet's Parallel and Recurrent Representation}
To better understand the connection between parallel and recurrent equations, consider the following matrix operations:
\label{sec:parallel_recurrent}

\small %

1. \textbf{Defining \(Q\), \(K^\top\), and \(V\):}
\[
Q =
\begin{bmatrix}
q_1 \\ q_2 \\ q_3
\end{bmatrix},
\quad
K^\top =
\begin{bmatrix}
k_1^\top & k_2^\top & k_3^\top
\end{bmatrix},
\quad
V =
\begin{bmatrix}
v_1 \\ v_2 \\ v_3
\end{bmatrix}.
\]

2. \textbf{Computing \(QK^\top\) and defining \(D\):}
\[
QK^\top =
\begin{bmatrix}
q_1 k_1^\top & q_1 k_2^\top & q_1 k_3^\top \\
q_2 k_1^\top & q_2 k_2^\top & q_2 k_3^\top \\
q_3 k_1^\top & q_3 k_2^\top & q_3 k_3^\top
\end{bmatrix},
\quad
D =
\begin{bmatrix}
1 & 0 & 0 \\
\gamma & 1 & 0 \\
\gamma^2 & \gamma & 1
\end{bmatrix}.
\]

3. \textbf{Computing the Retention Matrix:}
\[
(QK^\top) \odot D =
\begin{bmatrix}
q_1 k_1^\top & 0 & 0 \\
\gamma q_2 k_1^\top & q_2 k_2^\top & 0 \\
\gamma^2 q_3 k_1^\top & \gamma q_3 k_2^\top & q_3 k_3^\top
\end{bmatrix}.
\]

4. \textbf{Final Retention Operation:}
\begin{align*}
Y &=
\begin{bmatrix}
q_1 (k_1^\top v_1) \\
q_2 (\gamma k_1^\top v_1 + k_2^\top v_2) \\
q_3 (\gamma^2 k_1^\top v_1 + \gamma k_2^\top v_2 + k_3^\top v_3)
\end{bmatrix} \nonumber \\
&=
\begin{bmatrix}
q_1 S_1 \\
q_2 S_2 \\
q_3 S_3
\end{bmatrix}, \quad
\text{where } S_n = \gamma S_{n-1} + k_n^{\top} v_n.
\label{eq:app-parrec-Y}
\end{align*}

\normalsize %

This formulation demonstrates the validity and flexibility of the retention mechanism.

\appsection{Derivation of the Asymptotic Complexity for Each Inference Form}
\label{supp:Asymptotic}
\appsubsection{Vanilla Form of Transformer Total Computation}

\[
y_n = \operatorname{softmax}\left(\frac{1}{\sqrt{d_k}} Q_n K_n^{\top}\right) V_n
\]

Where the vector shapes are \(Q_n \in \mathbb{R}^{n\times d}\), \(K_n^{\top} \in \mathbb{R}^{d\times n}\), and \(V_n \in \mathbb{R}^{n\times d}\).

\begin{enumerate}
    \item[1.] Compute \(Q_n K_n^{\top}\)
    \begin{align*}
        Q_n \cdot K_n^{\top} &= \\ %
        \begin{bmatrix}
        q_{11} & q_{12} & \cdots & q_{1d} \\
        q_{21} & q_{22} & \cdots & q_{2d} \\
        \vdots & \vdots & \ddots & \vdots \\
        q_{n1} & q_{n2} & \cdots & q_{nd}
        \end{bmatrix}
        &\times
        \begin{bmatrix}
        k_{11} & k_{12} & \cdots & k_{1n} \\
        k_{21} & k_{22} & \cdots & k_{2n} \\
        \vdots & \vdots & \ddots & \vdots \\
        k_{d1} & k_{d2} & \cdots & k_{dn}
        \end{bmatrix}
    \end{align*}
    \begin{itemize}
        \item Total Computations: \(n \times d \times n = n^2 \times d\) multiplications, \(n^2 - 1\) additions.
    \end{itemize}

    \item[2.] Compute softmax \(\left(\frac{1}{\sqrt{d_k}} Q_n K_n^{\top} \cdot \text{Mask} \right) \times V_n\):
    \begin{align*}
        \operatorname{attn} \times V_n &= \\ %
        \begin{bmatrix}
        \alpha_{11} & \alpha_{12} & \cdots & \alpha_{1n} \\
        \alpha_{21} & \alpha_{22} & \cdots & \alpha_{2n} \\
        \vdots & \vdots & \ddots & \vdots \\
        \alpha_{n1} & \alpha_{n2} & \cdots & \alpha_{nn}
        \end{bmatrix}
        &\times
        \begin{bmatrix}
        v_{n11} & v_{n12} & \cdots & v_{n1d} \\
        v_{n21} & v_{n22} & \cdots & v_{n2d} \\
        \vdots & \vdots & \ddots & \vdots \\
        v_{nn1} & v_{nn2} & \cdots & v_{nnd}
        \end{bmatrix}
    \end{align*}
    \begin{itemize}
        \item Total Computations: \(n \times d \times n = n^2 \times d\) multiplications, \(n \times (d-1)\) additions.
        \item Total Computations for Modified Form with Causal Mask:
        \[
        \left(2 n^2 \times d + n^2 - 1 + n \times (d-1)\right),
        \]
        which simplifies to \(O\left(n^2 \times d\right)\).
        \item This process involves matrix multiplications across all \(n\) tokens for each generation step, resulting in \(O\left(n^2 d\right)\) complexity per step.

    \end{itemize}
\end{enumerate}

The inefficiency is due to the causal nature of the decoder that redundantly recomputes the same attention weights for previously processed tokens.

\appsubsection{Key-Value (KV) Cached Form}

To address this inefficiency, instead of recomputing attention for all previous tokens, KV Caching stores the K and V from earlier steps. This modifies the computation as follows:

\begin{enumerate}
    \item[1] \textbf{Key-Value Cached Form Total Computation}:
    \[
    y_n = \operatorname{softmax}\left(\frac{1}{\sqrt{d_k}} q_n\left[K_{n-1}^{\top} \mid k_n^{\top}\right]\right)\left[\begin{array}{c}
    V_{n-1} \\
    v_n
    \end{array}\right]
    \]

    \begin{itemize}
      \item Vector shapes: $q_n:(1,d)$,
      \[
        \left[K_{n-1}^{\top}\mid k_n^{\top}\right]:(d,n),\quad
        \left[\begin{array}{c}V_{n-1}\\ v_n\end{array}\right]:(n,d)
      \]
    \end{itemize}

    \item[2.] \textbf{Detailed Computation}:
    \begin{itemize}
        \item Equation: \(q_n\left[K_{n-1}^{\top} \mid k_n^{\top}\right] = q_n K_n^{\top}\)
        \item Expanded as:
        \begin{align*}
            q_n \cdot
            \left[
            \begin{array}{llll}
                K_{n-1}^{\top} \mid k_n^{\top}
            \end{array}
            \right]  =
            \\ %
            \begin{bmatrix}
                q_{n1} & q_{n2} & \cdots & q_{nd}
            \end{bmatrix}
            &\times
            \begin{bmatrix}
                k_{11} & k_{12} & \cdots & k_{1n} \\
                k_{21} & k_{22} & \cdots & k_{2n} \\
                \vdots & \vdots & \ddots & \vdots \\
                k_{d1} & k_{d2} & \cdots & k_{dn}
            \end{bmatrix}
        \end{align*}

        \item Total Computations: \(d \times n\) multiplications, \(n-1\) additions.
    \end{itemize}

    \item[3.] \textbf{Output Computation}:
    \[
    \operatorname{softmax}\left(\frac{1}{\sqrt{d_k}} q_n\left[K_{n-1}^{\top} \mid k_n^{\top}\right]\right)
    \left[
    \begin{array}{c}
    V_{n-1} \\
    v_n
    \end{array}
    \right]
    =
    \operatorname{attn} \times V_n
    \]

    \begin{itemize}
        \item Expanded as:
        \begin{align*}
            \operatorname{attn} \times V_n &= \\ %
            \begin{bmatrix}
                \alpha_1 & \alpha_2 & \cdots & \alpha_n
            \end{bmatrix}
            &\times
            \begin{bmatrix}
                v_{n11} & v_{n12} & \cdots & v_{n1d} \\
                v_{n21} & v_{n22} & \cdots & v_{n2d} \\
                \vdots & \vdots & \ddots & \vdots \\
                v_{nn1} & v_{nn2} & \cdots & v_{nnd}
            \end{bmatrix}
        \end{align*}

        \item Total Computations: \(d \times n\) multiplications, \(n-1\) additions.
    \end{itemize}
\end{enumerate}

\begin{itemize}
    \item \textbf{Total Computations for Key-Value Cached Form:}
    \[
    2dn + 2(n-1), \text{ which is } O(nd).
    \]
    \item The overall computation is significantly reduced as it only computes the attention for the new token using cached keys and values.
\end{itemize}

\appsubsection{RetNet's Recurrent Representation}

RetNet can be considered a more advanced form of key-value caching. By discarding the softmax operation and introducing gamma decay, there is no longer a need to cache \(K\) and \(V\) separately. Instead, the network only needs to store the product \(k_n^{\top} v_n\) at each step and update a recurrent state representation \(S_n\) by adding these values.

\[
S_n = \gamma S_{n-1} + k_n^{\top} v_n
\]

\[
y_n = q_n S_n
\]

\begin{enumerate}
    \item[1.] \textbf{Equation:} \(S_n = \gamma S_{n-1} + k_n^{\top} v_n\)
    \begin{itemize}
        \item Matrix Multiplication: Simplifying to the last two dimensions \((d, 1) \cdot (1, d)\)
        \begin{align*}
            k_n^{\top} \cdot v_n &= \\ %
            &\begin{bmatrix}
                k_{n1} \\
                k_{n2} \\
                \vdots \\
                k_{nd}
            \end{bmatrix}
            \times
            \begin{bmatrix}
                v_{n1} & v_{n2} & \cdots & v_{nd}
            \end{bmatrix} \\ \\
            &=
            \begin{bmatrix}
                k_{n1}v_{n1} & k_{n1}v_{n2} & \cdots & k_{n1}v_{nd} \\
                k_{n2}v_{n1} & k_{n2}v_{n2} & \cdots & k_{n2}v_{nd} \\
                \vdots & \vdots & \ddots & \vdots \\
                k_{nd}v_{n1} & k_{nd}v_{n2} & \cdots & k_{nd}v_{nd}
            \end{bmatrix}
        \end{align*}
        \item Total Computations: \(d^2\) multiplications, 0 additions.
    \end{itemize}

    \item[2.] \textbf{Equation:} \(y_n = q_n S_n\)
    \begin{itemize}
        \item Matrix Multiplication: Simplifying to the last two dimensions \((1, d) \cdot (d, d)\)
\begin{adjustwidth}{-\leftmargin}{0pt}
\begin{align*}
& \quad \quad \quad \quad  y_n = q_n \cdot S_n \\
&= \begin{bmatrix}
    q_{n1} & q_{n2} & \cdots & q_{nd}
  \end{bmatrix}
  \times
  \begin{bmatrix}
    S_{n11} & S_{n12} & \cdots & S_{n1d} \\
    S_{n21} & S_{n22} & \cdots & S_{n2d} \\
    \vdots  & \vdots  & \ddots & \vdots \\
    S_{nd1} & S_{nd2} & \cdots & S_{ndd}
  \end{bmatrix} \\
&\quad \quad \quad \quad \quad  =  \begin{bmatrix}
    y_{n1} & y_{n2} & \cdots & y_{nd}
  \end{bmatrix}
\end{align*}
\end{adjustwidth}
\item Total Computations: \(d^2\) multiplications, \(d-1\) additions.
    \end{itemize}
\end{enumerate}

\begin{itemize}
    \item \textbf{Total Computations for Recurrent Form:}
    \[
    2d^2 + d - 1, \text{ which is } O(d^2).
    \]
\end{itemize}

\appsection{Beam Search Decoding Memory Complexity}
\label{supp:beam_m}

\appsubsection{KV-Cached Decoding}

\smallskip
\textbf{Persistent Cache Storage:} \\
For each layer, we store the keys and values for all decoded text tokens. Their typical shape is:
\begin{equation*}
\text{Keys (or Values)} : \quad (B, H, N, d_h)
\end{equation*}

Thus, the memory required per cache (key or value) is:

\begin{equation*}
M_{\text{persistent, per cache}} = B \times H \times N \times d_h = B \times N \times d
\end{equation*}

Since both keys and values are stored, the total persistent memory is:

\begin{equation*}
M_{\text{persistent}}^{\mathrm{KV}} = 2 B N d
\end{equation*}

\smallskip
\noindent \textbf{Temporary Allocations during Concatenation:} \\
At each decoding step \( t \) (where \( t \leq N \)), the model concatenates the new key and value
\( (B, H, 1, d_h) \) with the previous cache \( (B, H, t-1, d_h) \), resulting in a new cache of shape \( (B, H, t, d_h) \).

The size of the temporary tensor created here for the concatenation is:

\begin{equation*}
B \times H \times t \times d_h = B \times t \times d
\end{equation*}

In the worst case (at \( t = N \)), the temporary allocation per cache is:

\begin{equation*}
M_{\text{temp, per cache}} \approx B \times N \times d
\end{equation*}

The total temporary memory required for both keys and values is:

\begin{equation*}
M_{\mathrm{temp}}^{\mathrm{KV}} \approx 2 B N d
\end{equation*}

\smallskip
\noindent \textbf{Peak Memory Usage for KV-Cached Decoding:} \\
The total peak memory, both persistent and temporary, can be expressed as:
\begin{equation*}
M_{\mathrm{KV}} \approx M_{\text{persistent}}^{\mathrm{KV}} + M_{\mathrm{temp}}^{\mathrm{KV}}
\approx 2 B N d + 2 B N d = 4 B N d
\end{equation*}

\appsubsection{Recurrent Decoding (RetNet Form)}

Here, instead of storing the entire history, a fixed-size recurrent state is maintained. For each layer, assume the state has the shape:

\begin{equation*}
(B, H, d_h, d_h)
\end{equation*}

Thus, the memory per layer is:

\begin{equation*}
M_{\text{Recurrent}} \approx B \times H \times d_h \times d_h = B \times H \times \left(\frac{d}{H}\right)^2 = B \times \frac{d^2}{H}
\end{equation*}

Note that this memory usage does not grow with \(N\).

\end{document}

%% file: acronyms.tex
\newacro{ctc}[CTC]{Connectionist Temporal Classification}
\newacro{hmm}[HMM]{Hidden Markov Model}
\newacro{htr}[HTR]{Handwritten Text Recognition}
\newacro{knn}[k-NN]{k-Nearest Neighbors}
\newacro{svm}[SVM]{support vector machines}
\newacro{cnn}[CNN]{Convolutional Neural Network}
\newacro{retnet}[RetNet]{Retentive Network}
\newacro{rnn}[RNN]{Recurrent Neural Network}
\newacro{bilstm}[BiLSTM]{Bidirectional Long Short-Term Memory}
\newacro{htr}[HTR]{Handwritten Text Recognition}

\newacro{armf}[ARMF]{Attention-Retention Modality Fusion}
\newacro{dtrhtr}[DTrHTR]{Decoder-only Transformer for HTR}
\newacro{kv}[KV]{Key-Value}
\newacro{pff}[PFF]{Position-wise Feed-Forward Network}
\newacro{mha}[MHA]{Multi-Head Attention}
\newacro{msr}[MSR]{Multi-Scale Retention}